\documentclass[11pt, letterpaper]{nyu}

\usepackage[authoryear,sort&compress,round]{natbib}
\usepackage[capitalize]{cleveref}

\newcommand{\caplong}{Contact-Anchored Policies}
\newcommand{\capshort}{CAP}
\newcommand{\gym}{EgoGym}



\title{Contact-Anchored Policies: Contact Conditioning Creates Strong Robot Utility Models}
\correspondingauthor{ Zichen Jeff Cui {<\href{mailto:jeff.cui@nyu.edu}{jeff.cui@nyu.edu}>}}
\reportnumber{}
\websitedef{\href{https://cap-policy.github.io}{cap-policy.github.io}}

\author[1]{Zichen Jeff Cui}
\author[3]{Omar Rayyan}
\author[2]{Haritheja Etukuru}
\author[1]{Bowen Tan}
\author[1]{Zavier Andrianarivo}
\author[1]{Zicheng Teng}
\author[1]{Yihang Zhou}
\author[6]{Krish Mehta}
\author[1]{Nicholas Wojno}
\author[1]{Kevin Yuanbo Wu}
\author[1]{Manan H Anjaria}
\author[1]{Ziyuan Wu}
\author[1]{Manrong Mao}
\author[1]{Guangxun Zhang}
\author[4]{Binit Shah}
\author[5]{Yejin Kim}
\author[1]{Soumith Chintala}
\author[1]{Lerrel Pinto}
\author[2]{Nur Muhammad Mahi Shafiullah}

\affil[1]{New York University,
    \textsuperscript{2}University of California, Berkeley,
    \textsuperscript{3}University of California, Los Angeles \protect\\ %
    \textsuperscript{4}Hello Robot Inc.,
    \textsuperscript{5}Ai2,
    \textsuperscript{6}University of Waterloo
}

\teaser{%
\begin{center}
\includegraphics[width=0.975\textwidth]{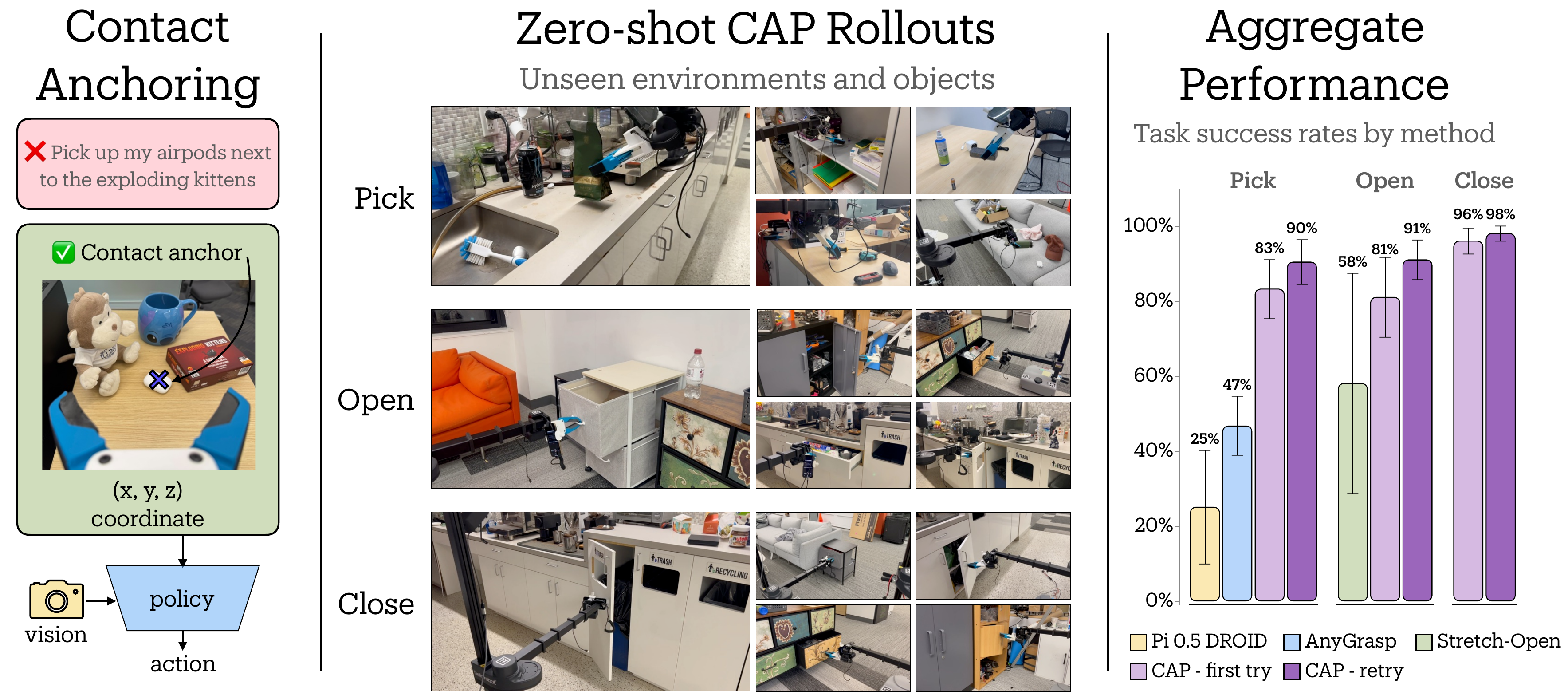}
\vspace{0.5em}
\captionof{figure}{We introduce~\caplong~(\capshort), a method to conditioning multimodal policies with physical contact information. Such policies are able to generalize zero-shot to novel objects and scenes with orders of magnitude less data, compute, and model parameters compared to frontier behavior model, while outperforming them on atomic skills trained with~\capshort{}.}
\label{fig:intro}
\end{center}
}

\begin{abstract}
The prevalent paradigm in robot learning attempts to generalize across environments, embodiments, and tasks with language prompts at runtime. A fundamental tension limits this approach: language is often too abstract to guide the concrete physical understanding required for robust manipulation. In this work, we introduce \textit{Contact-Anchored Policies} (\capshort{}), which replace language conditioning with points of physical contact in space. Simultaneously, we structure \capshort{} as a library of modular utility models rather than a monolithic generalist policy. This factorization allows us to implement a real-to-sim iteration cycle: we build EgoGym, a lightweight simulation benchmark, to rapidly identify failure modes and refine our models and datasets prior to real-world deployment. We show that by conditioning on contact and iterating via simulation, \capshort{} generalizes to novel environments and embodiments out of the box on three fundamental manipulation skills while using only 23 hours of demonstration data, and outperforms large, state-of-the-art VLAs in zero-shot evaluations by 56\%. All model checkpoints, codebase, hardware, simulation, and datasets will be open-sourced.

\end{abstract}

\begin{document}

\maketitle

\section{Introduction}

Now is the age for general robots.
Yet, the resource requirements of training general policies is burgeoning constantly. Today it is measured in thousands: of human data collection hours, GPU cluster size, and numbers of real world evaluations.
And even with all the resources, their generalization abilities remain more limited than a young child or household pet.
What is the cause behind such a stark gap?
One probable cause is that our current pipeline of learning general physical behavior may be backwards.
Even though language is only a recent acquisition on our evolutionary ladder of skills, many of our current general robot policies are built on top of large language model bases.
Folk wisdom holds that this internet-trained language backbone is necessary for generalization because language conditioning is how diverse behavior is elicited from the model.
However, language is as a medium for information for robot suffers from a few critical problems. First, language is imprecise: robotics needs precise spatial awareness which is not easy to convey in natural language abstractions. Second, language understanding comes at a cost of increasingly large model size leading to inefficient inference. These models are full of extraneous information, like distance between the earth and moon, that may be entirely unnecessary for a general robot.

In this work, we propose a simple fix: instead of natural language, we propose \textit{physical contacts} as the policy medium.
Instead of modeling the robot observation and action together with an underspecified language description of the task, we model observation and action jointly with physical contact the robot makes with the environment.
We call such policies \caplong{}~(\capshort{}).
With our simple change, we are able to train general policies for three set of common activities: picking up objects, and opening and closing doors and drawers; all from only 23 hours of human demonstrations.
On zero-shot evaluations in fully novel scenes and objects, our models outperform state-of-the-art generalist vision-language-action models such as $\pi_{0.5}$~\citep{PI2025Pi05}.
Moreover, since our policies are trained on handheld gripper data, we are able to deploy our policies on multiple robot embodiments out of the box.

Developing such policies efficiently also requires multiple iterations on modeling and dataset curation. Typically, such iterations require training and evaluating models on the target set of tasks.
To take advantage of the facts that (a) our target tasks are factored and (b) our primary goal is zero-shot environment and object generalization, we develop a lightweight simulation benchmark, EgoGym, as our key iteration metric.
This benchmark focuses primarily on object and scene diversity and trades off photorealism for speed.
As our primary goal is generalization, we find that success in these simulation environments under distribution shift is a great metric for capturing the emergence of general behavior.
\vspace{-1em}

\begin{figure}[t!]
    \centering
    \includegraphics[width=1\linewidth]{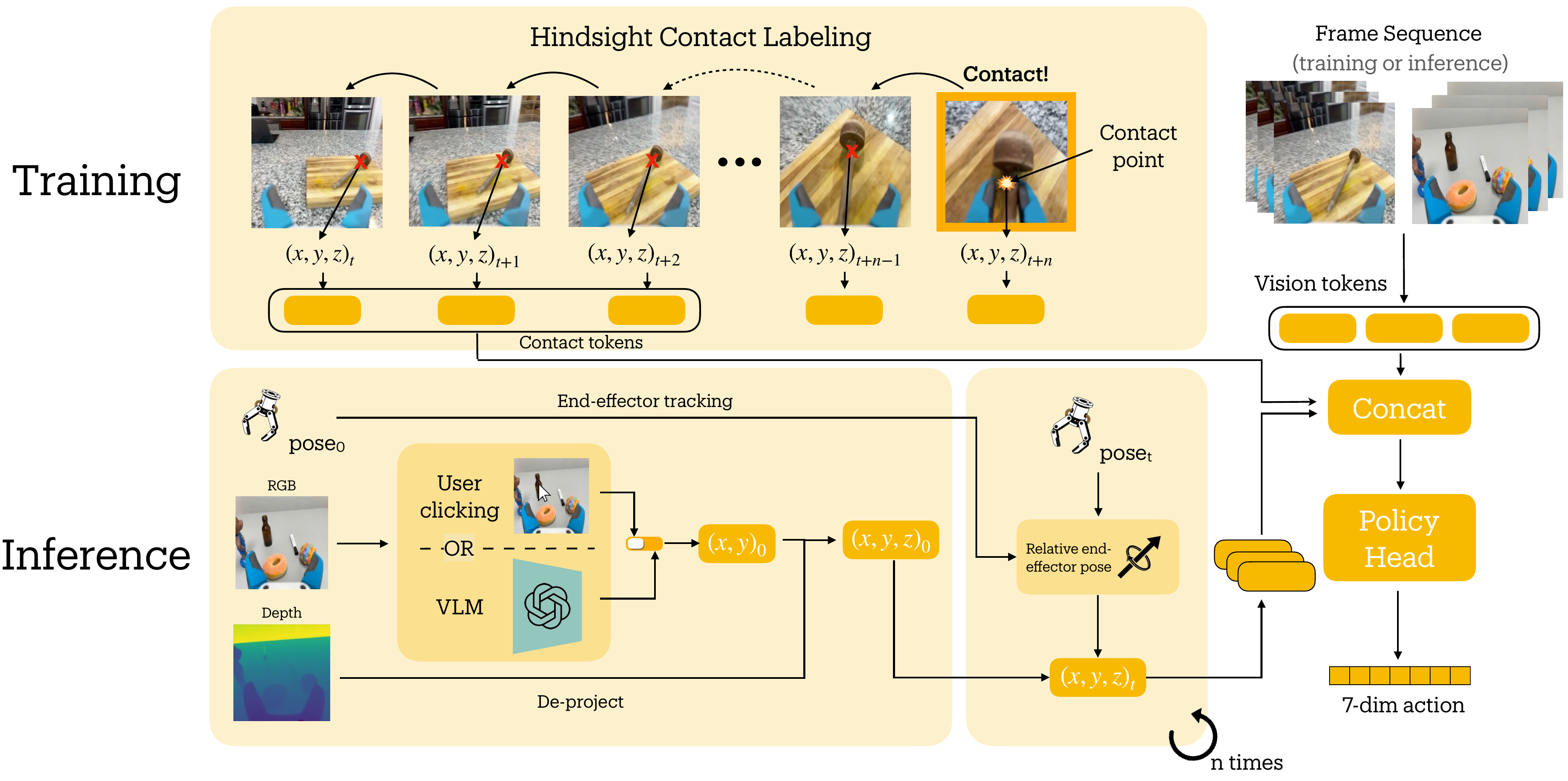}
    \caption{The process of data labeling, training, and inference for~\caplong{}. (a) During training, we detect the contact point from the data and label the trajectory with hindsight relabeling. (b) During inference, we use a user click or VLM conditioned on user command to derive the contact condition. In both cases, the contact tokens and visual tokens get concatenated and passed to the model which uses them as input to predict the actions.}
    \label{fig:overview}
\end{figure}
\section{Background}
\paragraph{Behavior Cloning} Behavior cloning (BC) is one of the primary ways of teaching robots intelligent behavior from humans. BC casts the problem of learning a robotic behavior policy $\pi$ that maps robotic observations $o \in \mathcal O$ to robotic actions $a \in \mathcal A$ as a supervised learning problem. Given a dataset $\mathcal D \subset \mathcal O \times \mathcal A$ of human demonstrations trajectories, BC defines a policy class $\Pi$ and an loss function $\mathcal L$ and trains a policy $\pi \in \Pi$ that minimizes the loss function $\mathcal L (\mathcal D)$. There has been significant research on finding the best learning objective~\citep{florence2021implicit,shafiullah2022behavior,chi2023diffusion,lee2024behavior} as well as methods for collecting BC datasets, such as leader-follower teleoperation~\citep{zhao2023aloha,wu2023gello}, VR teleop~\citep{iyer2024open,cheng2024opentelevisionteleoperationimmersiveactive}, and handheld tools~\citep{song2020grasping,shafiullah2023bringing,chi2024universal}.
\paragraph{Vector Quantized Behavior Transformer (VQ-BeT)} VQ-BeT~\citep{lee2024behavior} is a behavior cloning algorithm designed to learn robotic behaviors from large, multi-modal behavior datasets. VQ-BeT is a two-stage algorithm. The first stage finds a self-supervised discrete action representation using the actions from the dataset by training a Residual Vector Quantized Variational Autoencoder (VQ-VAE). Then, second stage trains an autoregressive transformer to predict the tokenized actions given the observation sequence. In this work, we use VQ-BeT for our behavior modeling as autoregressive architectures are more straightfoward to condition compared to diffusion models~\citep{chi2023diffusion,tri2025lbm}, and they lead to smaller and faster models.
\paragraph{Robot Utility Models} The advent of large robotic datasets~\citep{shafiullah2023bringing,padalkar2023open,khazatsky2024droid} has training policies that can generalize to novel environments, objects, or tasks. Many large, proprietary models focus on task-level generalization~\citep{Black2024Pi0,PI2025Pi05,NVIDIA2025GR00T} as the primary generalization axis. Our work is inspired by~\citet{etukuru2024rum} which showed that performing a single task generally in diverse scenes and robots can be done efficiently, with a few hours of demonstrations, with the diverse, high quality data and a multi-modal behavior cloning policy. This work also introduces verifier-guided retrying for robotics, where with guidance from an automated verified, a robot gets to retry a task until it is stuck or successful.

\section{Contact-Anchored Policies}

\subsection{Data Collection and Contact Annotations}

\begin{figure}[h!]
    \centering
    \includegraphics[width=0.55\linewidth]{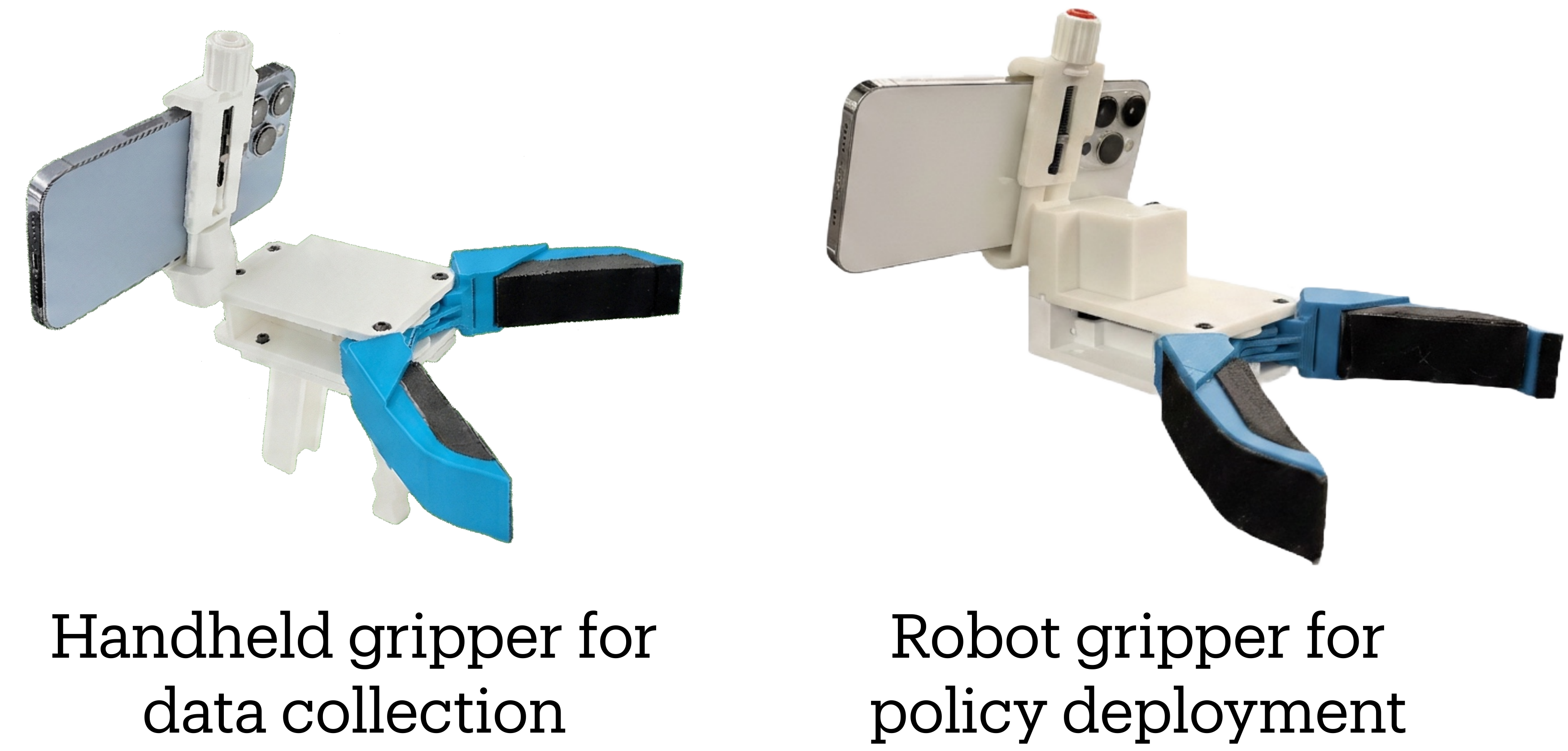}
    \caption{Our data collection tool and matching robot deployment gripper.}
    \label{fig:gripper}
\end{figure}
\subsubsection{Gripper Hardware Design}
To minimize the embodiment gap between data collection and robot inference, we design a low-cost, 3D-printable gripper compatible with both handheld operation and robot mounting. The handheld gripper is designed to be lightweight, ergonomic, convenient, and strong. Apart from the iPhone, the gripper itself consists almost entirely of 3D printed parts. We’ve designed the trigger handle such that the closing mechanism feels natural, allowing for prolonged use. Its small form factor makes it easy to throw into a backpack, and collect data in any location. The gripper design comprises an angular jaw 2-fingered mechanism, that allows for greater force and pinching of small objects. Similarly, the robot end-effector features compliant, back-drivable fingers with deformable foam padding for stable grasping across a diverse range of rigid and deformable objects. We use an iPhone 13 Pro as the primary sensor suite for both data collection and inference. The phone is rigidly mounted to the gripper chassis. For data collection, the gripper fingers are actuated by a handle. For inference, the same gripper modules are driven by a Dynamixel XL430 servo. This unified design ensures that the observation space remains consistent between the expert demonstrations and the robot's policy execution.

\subsubsection{Data collection}
We collect expert demonstrations for three primary tasks: Pick, Open, and Close. Data is collected with the \textit{AnySense} iOS application~\citep{bhirangi2024anysense}. The app records synchronized RGB-D streams and 6-DoF camera poses via ARKit visual-inertial odometry at 30Hz. Following \citet{etukuru2024rum}, we prioritize collecting data in diverse environments with varying lighting conditions, background clutter, and task object form. Trajectories with tracking loss, significant jitter, or failed task completions are discarded. The final dataset contains 20${,}$365 demonstrations (23.1 hours) across 424 environments, consisting of:
\begin{itemize}[leftmargin=*]
    \item \textbf{Pick:} general object pickup; 14${,}$606 demonstrations (16.3 hours) across 289 environments.
    \item \textbf{Open:} opening drawers and cabinet doors; 3${,}$690 demonstrations (4.7 hours) across 87 environments.
    \item \textbf{Close:} closing drawers and cabinet doors; 2${,}$069 demonstrations (2.0 hours) across 48 environments.
\end{itemize}

\subsubsection{Data preprocessing}
\textbf{Observation}: we resize RGB and depth images to $224 \times 224$, and augment the data with horizontal flip on the RGB-D observations and the corresponding camera odometry. We find that this helps the policy generalize to left-right symmetries such as cabinet doors. \textbf{Action}: the action space consists of the delta end-effector (EE) pose and the gripper aperture. Since we mount the iPhone rigidly onto the gripper, we can extract the delta EE pose directly from the iPhone camera pose trajectory. \textbf{Visual gripper state estimation}: we extract gripper action labels directly from the visual observations using SAM2~\citep{ravi2024sam2}. For each frame, we segment the left and right fingers and compute the centroid of their respective masks. The distance between these centroids is linearly mapped to a scalar $\in [0, 1]$, representing fully closed to fully open. Further details are in App. \cref{app:sec:data_collection}.

\subsubsection{Hindsight contact labeling}
We define the \textit{Contact Anchor} as a 3D coordinate $p$ where the policy is expected to interact with the object. To generate these labels for training, we do the following steps (illustrated in \cref{fig:overview}):
\begin{itemize}[leftmargin=*]
    \item \textbf{Contact Detection:} We first identify the timestep of contact, $t = c$. For Pick and Open tasks, this is naturally defined as the frame where the gripper aperture ceases to decrease, signaling that the fingers have made physical contact and halted against the object geometry. For the Close task, we label the contact frame during data collection by closing the grippers upon contact with the door.
    \item \textbf{Anchor Definition:} At $t = c$, we instantiate the contact anchor $p_c$ as the 3D coordinate centered between the gripper fingers in the camera frame.
    \item \textbf{Anchor Propagation:} For all previous timesteps $t < t_c$, we generate contact anchors with hindsight relabeling by back-projecting $p_c$ using the recorded camera odometry. Let $A_t \in \mathrm{SE}(3)$ denote the camera pose in the world frame at timestep $t$. The contact anchor in the camera frame for all previous timesteps $t < t_c$ is then simply given by $p_t = A_t^{-1} A_c p_c$ as the gripper approaches the contact. For the remainder of the episode $t > t_c$, in tasks such as Pick or Open, the object moves rigidly with the gripper as the gripper establishes contact. We freeze the anchor and simply repeat $p_c$ until the end of the episode.
\end{itemize}

\subsection{Policy Learning}
We formulate the policy learning as a conditional imitation learning problem, where the robotic policy $\pi(a_{t:t+h} | o_{t-k:t}, p_{t-k:t})$ predicts actions given a history of visual inputs and contact anchors. We implement this using a Vector-Quantized Behavior Transformer (VQ-BeT) \citep{lee2024behavior} with an observation context length of $k=3$. For the visual (RGB) observation, we pretrain a ResNet-50 backbone with MoCo~\citep{chen2021mocov3} on our dataset. For each timestep, we embed the $224 \times 224$ RGB input into a feature vector $z_v \in \mathbb{R}^{256}$. Separately, the contact anchor $p_t \in \mathbb{R}^3$, expressed in the current camera frame, is linearly projected to a contact embedding $z_c \in \mathbb{R}^{256}$. We concatenate these embeddings to form the observation token $s_t = [z_v, z_c]$. We feed a context window of observation tokens $s_{t-k:t}$ into VQ-BeT to predict the demonstrated actions. Each action $a_t$ consists of the delta end-effector pose and the continuous gripper position command. By conditioning the action distribution jointly on the RGB observation and the contact anchor, the policy adapts to diverse object geometry while anchoring its manipulation trajectory to the intended interaction point.

\subsection{Contact Prompting during Inference}
Unlike training, where contact anchors are derived from hindsight, inference requires an initial anchor $p_0$ before execution.
Given the initial RGB-D observation, a pixel coordinate $(u, v)$ is selected on the object of interest. This selection can be performed manually, or by querying an off-the-shelf VLM (e.g. Gemini Robotics-ER 1.5~\citep{team2025gemini}) with a text prompt (``point to the red mug''). Then, we deproject the 2D pixel $(u, v)$ using the depth map value $d_{u, v}$ and camera intrinsics $K$ to obtain the initial contact anchor in the camera frame, $p_0 = d_{u, v}K^{-1}[u, v, 1]^T$. As the robot executes the policy, the camera frame moves with the gripper. We track the anchor in the camera frame using the robot's forward kinematics, which provides higher accuracy than visual-inertial odometry. Let $A_t \in \mathrm{SE}(3)$ be the camera pose in the world frame at time $t$, derived from the robot's kinematic chain; the anchor $p_t$ is simply updated via $p_t = A_t^{-1} A_0 p_0$, so that the policy sees a consistent contact anchor in the world frame (see \cref{fig:overview}). After the gripper closes, we freeze the contact anchor to match the training data distribution.

\subsection{Simulation-in-the-loop Development}

\begin{figure}[h]
    \centering
\includegraphics[width=0.65\linewidth]{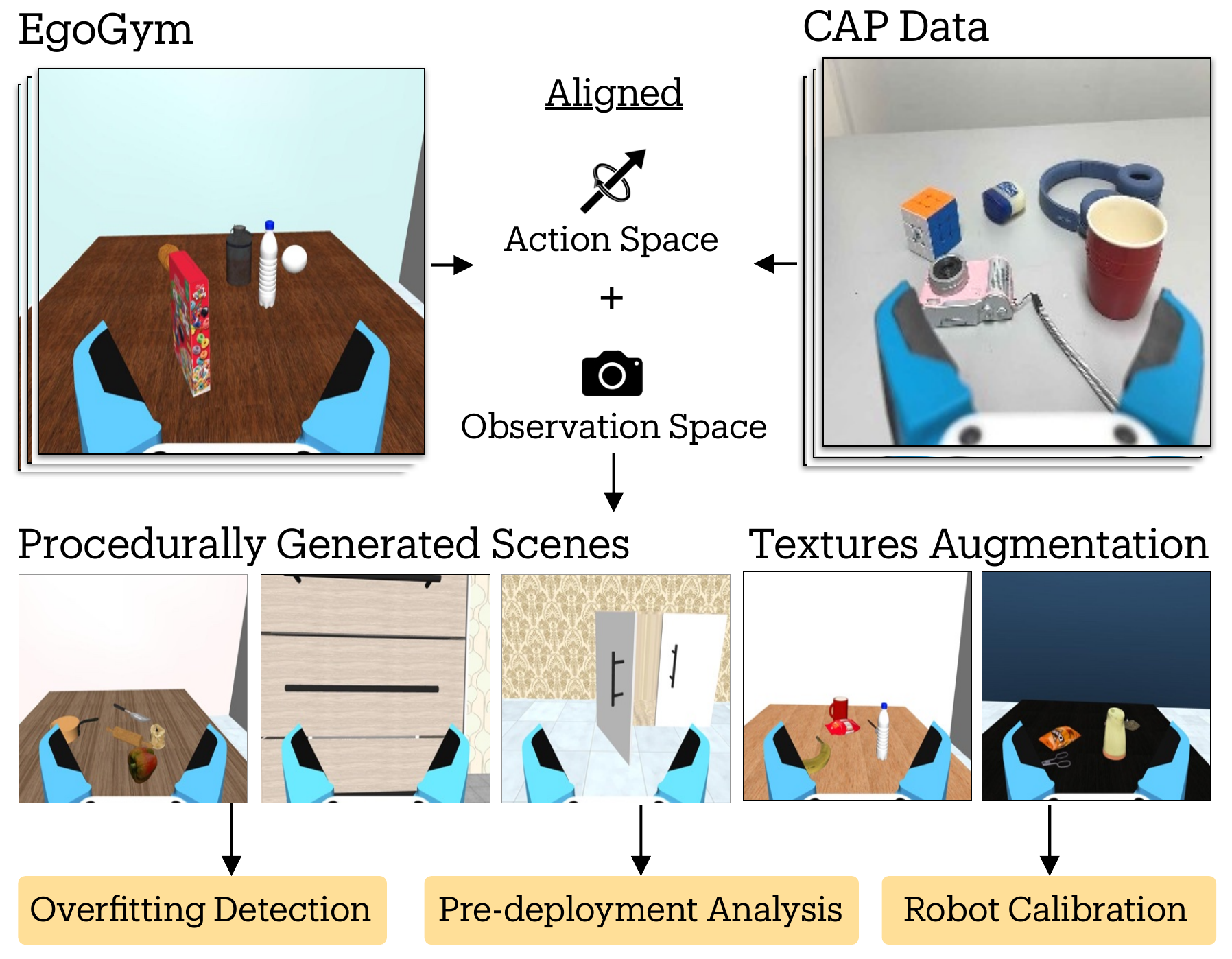}
\caption{EgoGym: a lightweight simulation-in-the-loop environment used for quick development and evaluation of \caplong{} (\capshort{}s). EgoGym enables fast checkpoint evaluation and failure mode discovery across Pick, Open, and Close tasks using procedurally generated scenes.}
    \label{fig:sim-scenes}
\end{figure}

To support rapid iteration and evaluation of CAPs, we develop EgoGym, a lightweight simulation suite used during policy training and development. It is designed with the motivation of (i) providing training signals beyond validation loss, which poorly correlates with real world performance, and (ii) accelerating CAP refinement by detecting failure modes prior to real-world deployment, and (iii) enabling the quick calibration of grasping thresholds we set for our policies.

EgoGym is implemented in MuJoCo \citep{todorov2012mujoco} and trades off visual realism in favor of scene diversity and execution speed. This allows it to run sufficiently fast to be included directly in the training loop of all 3 of our policies, enabling frequent checkpoint evaluation to detect overfitting.  We induce diversity through task-specific procedural scene generation. For our pick task, objects are sampled from a pool of 915 Objaverse~\citep{deitke2023objaverse} assets and spawned with varying poses and arrangements. For our opening and closing tasks, we procedurally generate articulated doors and drawers with randomized geometrical parameters at run-time. Across all three tasks, additional diversity is introduced by randomizing surface textures and adding distractor objects. Additional details on the assets and the randomization employed are provided in App.~\cref{app:sec:simulation}.

\section{Evaluating \capshort{}}
\begin{figure}[ht!]
    \centering
    \includegraphics[width=0.85\linewidth]{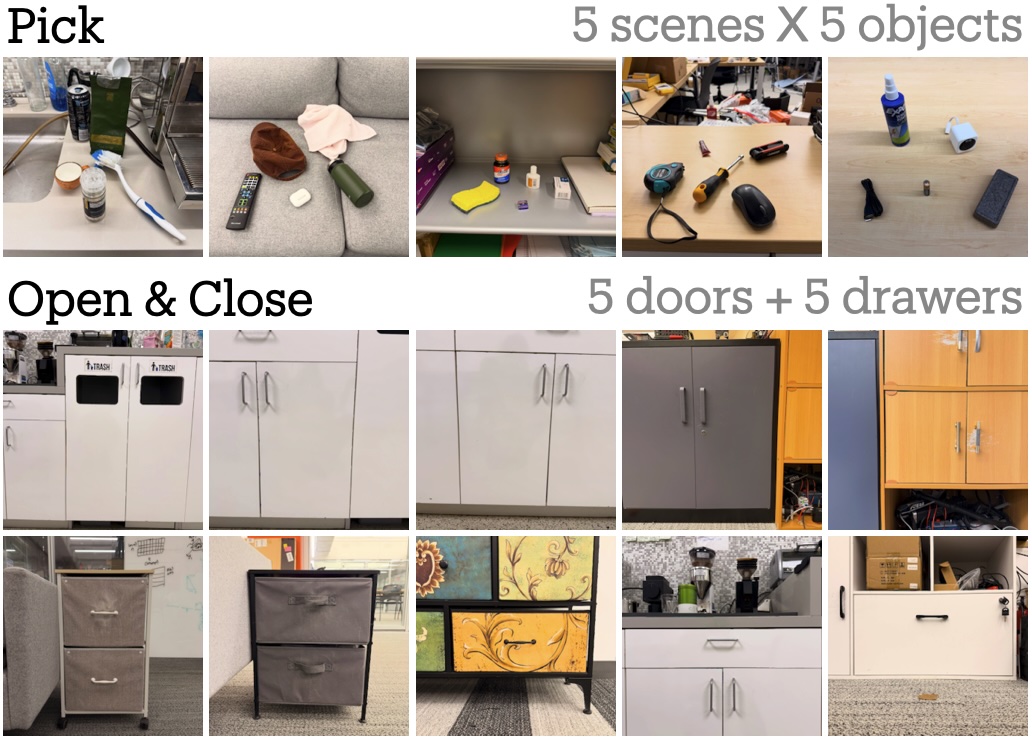}
    \caption{Evaluation environments for~\capshort{}. Each scene and object combination has 10 trials, so Pick checkpoints are evaluated for 250 episodes and Open or Close checkpoints are evaluated for 100 episodes.}
    \label{fig:eval_envs}
\end{figure}
We evaluate the zero-shot generalization performance of \caplong{} in diverse real and simulated environments, across multiple robot embodiments, including independent third-party evaluations at three external institutions. Our experiments are designed to answer the following questions:
\begin{enumerate}
\item How well does \capshort{} generalize zero-shot to unseen environments and objects?
\item Can \capshort{} work across robot embodiments out of the box?
\item Can \capshort{} be used to compose long-horizon manipulation behavior with tool calling?
\item Do simulated evaluations of \capshort{} reflect its real-world performance, and if so, can we use it to get detailed understanding of our policy performance?
\end{enumerate}

\subsection{Zero-shot Environment Generalization}
We evaluate \capshort{} on the three core manipulation tasks represented in our dataset: Pick, Open, and Close. All evaluations are zero-shot on unseen environments with no additional fine-tuning. We manually provide the policy with oracle contact prompts to isolate the generalization performance of \capshort{}.

\subsubsection{Pick} \label{ssec:pick-eval-main} We evaluate \capshort{} in five unseen scenes (kitchen, couch, meeting room, storage cabinet, work area) on the Stretch 3 platform. For each scene, we present the policy with a set of five objects not seen during training, for total 25 objects (\cref{fig:eval_envs}). The policy attempts to pick up each of the objects for 10 trials, randomizing the robot's initial position by $16\times11$cm (horizontal $\times$ vertical), for 250 total trials.

\subsubsection{Open \& Close} \label{ssec:open-close-eval-main} We evaluate \capshort{} on five cabinet doors and five drawers unseen during training (\cref{fig:eval_envs}), on Stretch 3. The policy attempts to open and close each door and drawer for 10 trials, randomizing the robot's initial position by $16\times11$cm (horizontal $\times$ vertical), for 100 total trials.
\begin{table}[ht!]
\caption{Evaluation results of~\capshort{} and baselines on our three different tasks and four different robot embodiments.}
\centering
\begin{tabular}{@{}llll@{}}
  \toprule
  Task         & Robot       & Model                           & Success rate \\ \midrule
  Pick         & Stretch     & CAP + Retry                     & $90.4\% \pm 6.0\% $ \\
  Pick         & Stretch     & CAP                             & $83.2\% \pm 7.9\% $ \\
  Pick         & Stretch     & AnyGrasp                        & $46.7\% \pm 7.9\% $ \\ 
  Pick         & Franka      & CAP                             & $79.0\% \pm 10.9\%$ \\
  Pick         & Franka      & CAP-VLM                         & $81.0\% \pm 9.2\% $ \\
  Pick         & Franka      & $\pi_{0.5\texttt{\tiny-DROID}}$ & $25.0\% \pm 15.2\%$ \\
  Pick         & XArm        & CAP                             & $83.0\% \pm 17.9\%$ \\
  Pick         & UR3e        & CAP                             & $70.0\% \pm 15.2\%$ \\
  \midrule 
  Open         & Stretch     & CAP + Retry                     & $91.0\% \pm 5.3\% $ \\
  Open         & Stretch     & CAP                             & $81.0\% \pm 10.7\%$ \\
  Open         & Stretch     & Stretch-Open                    & $58.0\% \pm 29.3\%$ \\ 
  \midrule
  Close        & Stretch     & CAP + Retry                     & $98.0\% \pm 3.0\% $ \\
  Close        & Stretch     & CAP                             & $96.0\% \pm 3.5\% $ \\ 
  \midrule
  EgoGym-Pick  & Sim Gripper & CAP                             & $79.88\% \pm 1.1\%$ \\
  EgoGym-Pick  & Sim Franka  & $\pi_{0.5\texttt{\tiny-DROID}}$ & $20.9\% \pm 1.1\%$ \\
  \bottomrule
\end{tabular}
\label{tab:results_table}
\end{table}
With oracle contact prompts, \capshort{} generalizes zero-shot to unseen environments, achieving a single-try performance of 83\% in Pick, 81\% in Open, and 96\% in Close (\cref{tab:results_table}).

Next, to set up fully autonomous rollouts, we evaluate whether we can autonomously generate contact prompts comparably to a human oracle. To propose contact anchors autonomously, we use Gemini Robotics-ER 1.5 \citep{team2025gemini}. We provide the VLM with the initial observation, and prompt it to point to the task object for the contact anchor (see App.~\cref{app:sec:deployment} for details). We evaluate \capshort{} with the same procedure described above. With VLM-generated contact anchors, \capshort{} achieves a comparable single-try performance of 81\% in Pick, 80\% in Open, and 97\% in Close, validating that automatic, VLM-generated contact prompts work as well as the human oracle.

\begin{figure}[ht!]
    \centering
    \includegraphics[width=0.55\linewidth]{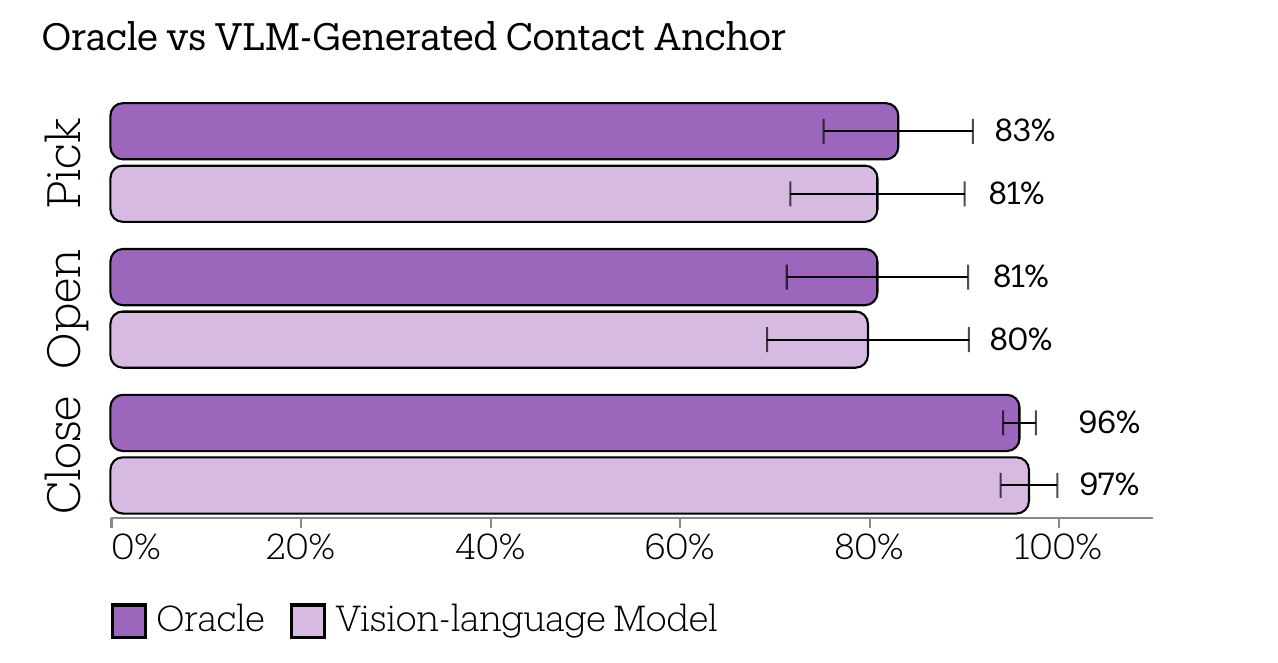}
    \caption{Comparison of contact instructions generated by human oracles and vision-language models for the~\capshort{}s. Downstream~\capshort{} performance is comparable on all tasks.}
    \label{fig:gt_vs_vlm}
\end{figure}

As we establish in~\cref{fig:gt_vs_vlm} that \capshort{} does not require a human in the loop, we add a VLM verifier for automatic retry upon failure. Following \citet{etukuru2024rum}, we use GPT-4o to verify whether the policy has successfully completed the task, using VLM-generated contact anchors for retries. We evaluate \capshort{} with the same criteria above, allowing up to 10 retries per trial. With automatic retries, \capshort{} achieves 90\% in Pick, 91\% in Open, and 98\% in Close (\cref{tab:results_table}, Fig. 1). The vast majority of failures are from verifier false positives, i.e. classifying a failed task as successfully completed.

\subsection{Zero-shot Embodiment Generalization}
\label{sec:xembodiment-evals}
\begin{figure}[th!]
    \centering
    \includegraphics[width=0.9\linewidth]{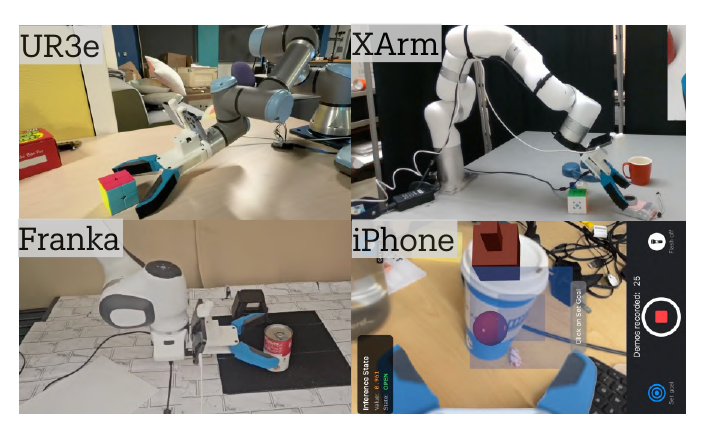}
    \caption{Cross-embodiment deployment of~\capshort{} on a Franka FR3, XArm 6, Universal Robotics UR3e, and an iPhone app. For the robots, the same~\capshort{} checkpoint generates EE-space motion that we translate to joint position control with IK. For the iPhone, the user is prompted to move the iPhone to where the robot should go next.}
    \label{fig:xemb_robots}
\end{figure}

\begin{figure}[ht!]
    \centering
    \includegraphics[width=0.55\linewidth]{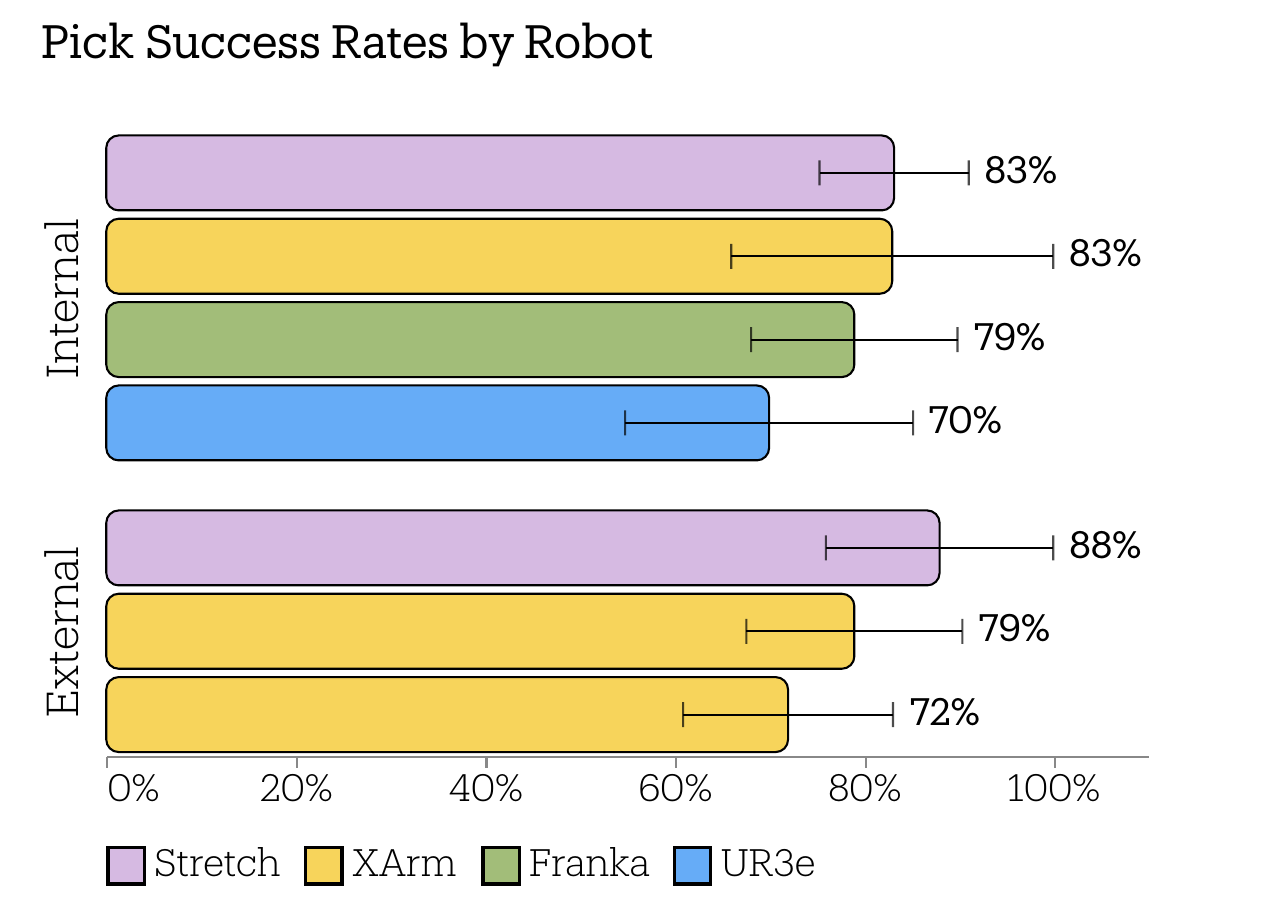}
    \caption{Evaluation of~\capshort{} zero-shot on diverse embodiments: each bar is a different set of evaluations in a unique site. To evaluate system robustness, we share checkpoints, setup instructions, and evaluation methodology to external collaborators and get performance numbers from them.}
    \label{fig:xembodiment}
\end{figure}

Since~\caplong{} are trained on handheld gripper data, they are theoretically compatible with any robot arm with six or more degrees of freedom. However, often because of idiosyncrasies in human behavior, policies trained with handheld tools have difficulty transferring to real robots by taking actions that violate robot kinematics. To validate cross-embodiment performance, beyond our primary embodiment in Stretch, we also evaluate our Pick policy on Franka FR3, XArm 6, and Universal Robotics UR3e (\cref{fig:xemb_robots}). For these evaluations, the same policy checkpoint is evaluated everywhere, and we only adapt our robot gripper mount and the inverse kinematic controller to the specific embodiments. Since these robots are mounted, we evaluate in the fixed environment with two sets of five unseen objects for a total of 10 objects, randomizing the object positions. The policy attempts to pick up each object for 10 trials, for a total of 100 trials. We see from~\cref{fig:xembodiment} that their success rates are comparable, showing the versatility of our policy. Out of the arms, UR3e achieves the lowest success because of its particularly short reach forward.

\subsubsection{External evaluations} Cross-embodiment evaluations of this kind are a test of system integration as much as they are of policy capabilities. We send our checkpoints and evaluation process to external collaborators at Hello Robot, UCLA, and Ai2 for independent evaluations. As we see in~\cref{fig:xembodiment}, result of such evaluations also broadly line up with internal evaluations.

\subsubsection{iPhone evaluations} Even before running a policy on the robot, it can be useful to understand if a policy will behave sensibly in a scene. Since our model has only 52 million parameters, we can deploy and infer from it in real time on modern iPhones. We develop an iPhone app that uses the camera as input stream, the ARKit API for pose tracking, and neural engine chips for inference. We rely on user taps for contact conditioning, and emulate a dummy robot gripper on screen to match observations. When a user taps on a target object, the app shows the next target position of the phone, as well as the predicted gripper motions. The app is able to show the user how~\capshort{} would navigate an in-the-wild scene and execute different grasps matching the target object affordance.

\subsection{Baseline Comparisons}

We compare \capshort{} with state-of-the-art task-specific and generalist baselines:
\begin{itemize}[leftmargin=*]
    \item $\pi_{0.5\texttt{\small-DROID}}$ \citep{PI2025Pi05} (Pick): a generalist VLA model trained on a large proprietary dataset and fine-tuned on DROID \citep{khazatsky2024droid}. We evaluate it on the DROID embodiment with a Robotiq 2F85 gripper mounted on a Franka FR3, a ZED Mini wrist camera, and a ZED 2i external camera. We use the prompt ``pick up the \texttt{\{object\}}'' and run the policy for 400 steps.
    \item AnyGrasp \citep{fang2023anygrasp} (Pick): an RGB-D grasp pose prediction model for object pickup that uses an external depth camera to generate grasps and then uses a planner to execute them. We evaluate it on the Hello Robot Stretch 3 robot platform.
    \item \texttt{stretch-open}~\citep{gupta2024opening} (Open): a modular pipeline for opening doors and drawers. The pipeline uses an RGB-D head camera to locate the handle, and generates and executes a motion plan to open it. We evaluate it on the Hello Robot Stretch 3 robot platform with the stock gripper to exactly match the embodiment in~\citep{gupta2024opening}.
\end{itemize}

We evaluate AnyGrasp with the same Stretch 3 evaluation procedure as described in \cref{ssec:pick-eval-main} for three trials per object, for a total of 75 trials. We evaluate $\pi_{0.5\texttt{\small-DROID}}$ with the Franka FR3 evaluation procedure as described in \cref{sec:xembodiment-evals}. For a fair comparison, we also evaluate \capshort{} with VLM-generated contact anchors with the same object description given to $\pi_{0.5\texttt{\small-DROID}}$, denoted as \capshort{}-VLM in \cref{tab:results_table}. AnyGrasp achieves a 47\% success rate, and $\pi_{0.5\texttt{\small-DROID}}$ achieves a 25\% success rate in our evaluations. \texttt{stretch-open} achieves a 58\% success rate. As shown in~\cref{tab:results_table}, \capshort{} outperforms comparable baselines by 23\% to 56\%.

\begin{figure}[t!]
    \centering
    \includegraphics[width=0.8\linewidth]{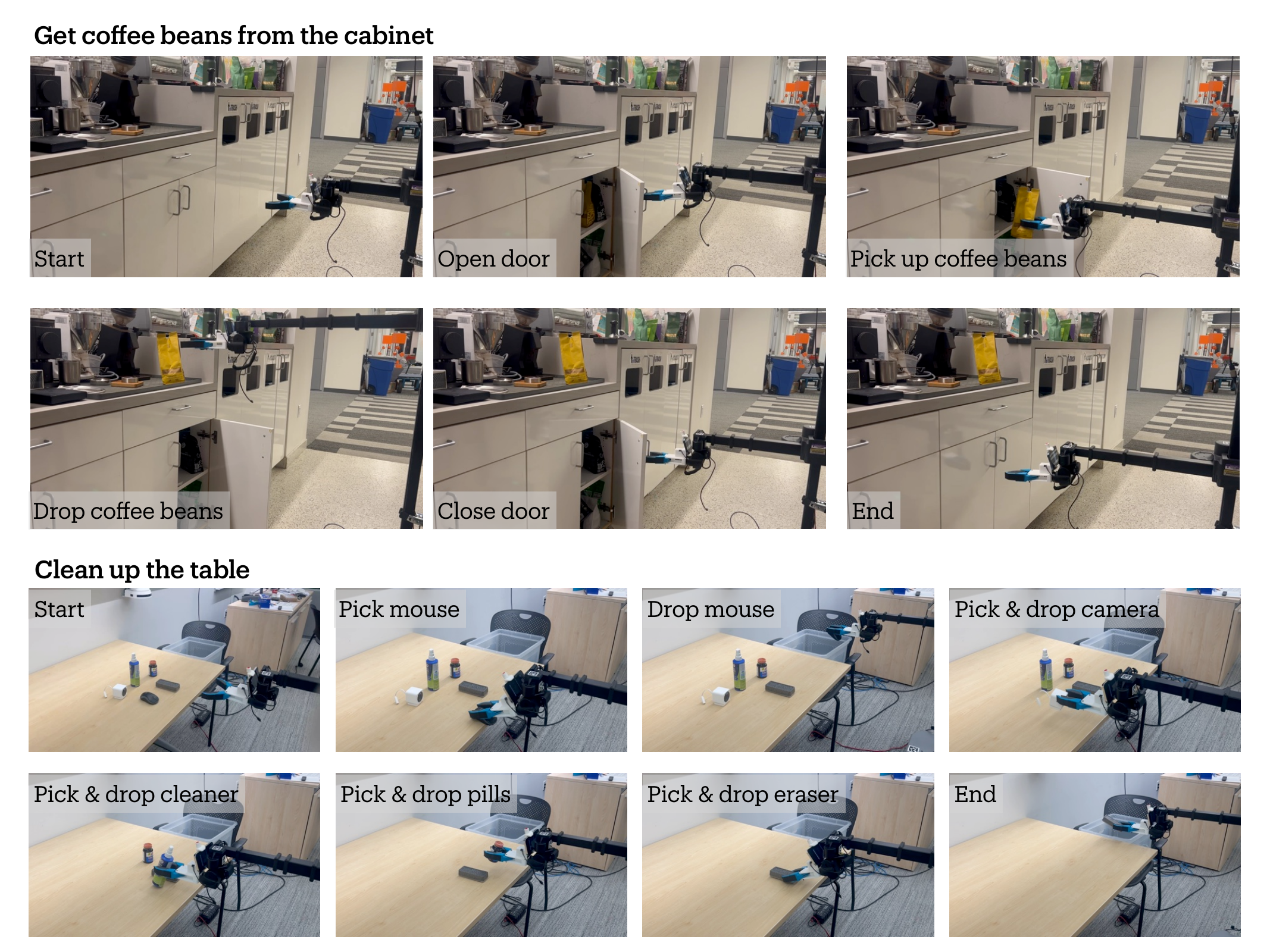}
    \caption{Performing long-horizon manipulations with~\caplong{} controlled by a high-level VLM controller via tool-calling. On the top, to retrieve coffee beans from cabinet, a controller combines Pick, Open, and Close~\capshort{}s, while on the bottom, a table cleanup is performed with Pick~\capshort{}.}
    \label{fig:long_horizon}
\end{figure}

\subsection{Eliciting Complex Behaviors with Tool Calling}
A standard argument for using large, end-to-end behavior policies is that theoretically they can exhibit long-horizon behavior that is unavailable to atomic utility models. However, because they are atomic, we posit that utility models can be chained together as a sequence of \textit{tool calls}~\citep{schick2023toolformer} by a larger model specializing in System 2 intelligence.

In~\cref{fig:gt_vs_vlm}, we verify that such models are able to generate contact conditions almost as well as a human for all three of our tasks. As a proof of concept, we take~\capshort{}s with verifier guided retrying and add a supervising high-level controller to get our robots to complete complex, long-horizon tasks as seen in~\cref{fig:long_horizon}.
We perform two long horizon tasks: fetching some coffee beans from a kitchen cabinet, and cleaning up a table. In the first task, the robot's goal is to retrieve a yellow bag of coffee beans from within the white kitchen cabinets. The robot needs to open the cabinet door, pick up the bag, drop it on the table, and close the door. In the second task, there are multiple objects on a table in front of the robot, and it has to move all of them into a bin next to it. The robot needs to perform a sequence of picks and drops until the table is clear. We use our trained Pick, Open, and Close policies, as well as scripts for executing ``Drop'' and ``Move the mobile base''. We run 10 experiments for each task.

\begin{table}[h!]
    \centering
    \caption{Success rate by stages for long-horizon tasks}
    \label{tab:composition-success}
    \begin{minipage}[t]{0.48\linewidth}
        \centering
        \begin{tabular}[t]{@{}lr@{}}
            \toprule
            \multicolumn{2}{c}{\textbf{Get coffee beans}} \\
            \midrule
            Stage & Success \\
            \midrule
            Open cabinet  & 10/10 \\
            Pick bag    & 7/10  \\
            Drop bag    & 7/10  \\
            Close cabinet & 6/10  \\
             &   \\
            \bottomrule
        \end{tabular}
    \end{minipage}
    \hfill
    \begin{minipage}[t]{0.48\linewidth}
        \centering
        \begin{tabular}[t]{@{}lr@{}}
            \toprule
            \multicolumn{2}{c}{\textbf{Clear table}} \\
            \midrule
            Stage & Success \\
            \midrule
            1st Object & 10/10 \\
            2nd Object & 10/10 \\
            3rd Object & 10/10 \\
            4th Object & 10/10 \\
            5th Object & 10/10 \\
            \bottomrule
        \end{tabular}
    \end{minipage}
\end{table}
Our compositional policy succeeds 6/10 trials on the coffee beans task, and 10/10 trials on the table cleaning task. \cref{tab:composition-success} describes the success rate by stages. For the coffee fetching task, most failures are from the policy opening the door partially, which the VLM verifier counts as a success, leading to hardware collisions with the door during the Pick stage. For the table cleaning task, the verifier can classify a successful grasp as a failure, leading to unnecessary retries, but the policy was able to move all objects into the bin for all 10 runs.

\subsection{Understanding Real Performance via Simulation}

\paragraph{Establishing Sim-and-real Correlation} 

\begin{figure}[h!]
    \centering
    \includegraphics[width=0.7\linewidth]{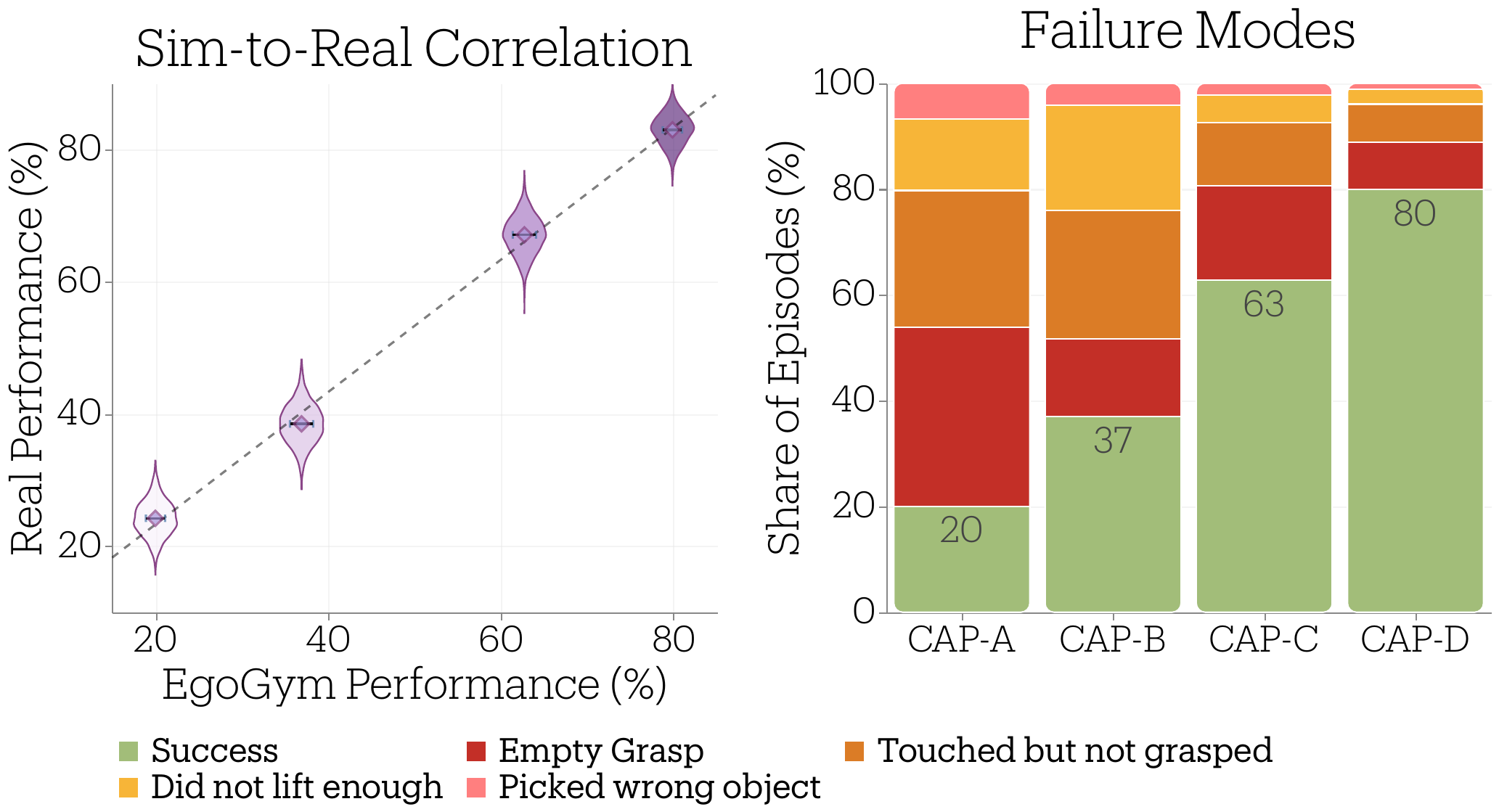}
    \caption{\textbf{Left:} Sim-to-real correlation for single-blind \gym{}-Pick evaluations. \textbf{Right:} Analysis of failure modes of four iterations of~\capshort{} Pick checkpoints in~\gym. With feedback from each iteration, our pipeline changes allowed better policy quality in real and sim.}
    \label{fig:sim2real&failures}
\end{figure}

To confirm whether CAPs behavior in \gym{} is indicative of real-world performance, we carry out a single-blind correlation study in which an evaluator, unaware of \gym{} results, is sent four Pick~\capshort{} checkpoints to evaluate in the real world. The evaluations includes 250 real-world runs and 5${\,}$000 \gym{} episodes, which include texture and objects augmentation as well as four random distractor objects per episode. While the real evaluation numbers are limited, the results in~\cref{fig:sim2real&failures} (left) suggest strong alignment between \gym{} and real-world performance across the checkpoints. For the rest of this section, we describe these four checkpoints with letters A through D in order of success rates.

\paragraph{Iterating CAP via failure analysis}

To better understand how policy behavior evolves across checkpoints, we analyze failure modes using \gym{} rollouts. Details of the failure categorization are provided in App.~\cref{app:sec:simulation}. \Cref{fig:sim2real&failures} (left) shows the distribution of outcomes over 5${\,}$000 simulated episodes for each analyzed checkpoint.

These observed failure modes motivated refinements to the CAP data processing pipeline. For example, the low-lift failures at checkpoint B revealed the existence of many post-grasp transitions with little or no end-effector motion, motivating us to introduce static-frame filtering which resulted in the following C checkpoint.

\subsection{Ablations}
We run ablation experiments to answer two questions related to~\caplong{}' performance:
\begin{itemize}[leftmargin=*]
    \item How important is the contact anchor? How does \capshort{} compare with an RGB-only policy?
    \item How does policy performance change as we increase distractor objects in the Pick task?
\end{itemize}

\subsubsection{Contact anchor ablation}
\begin{table}[h!]
\caption{Ablation results of~\capshort{} on the Close task.}
\centering
\begin{tabular}{@{}ll@{}}
  \toprule
  Model                           & Success rate \\
  \midrule
  CAP - RGB Only Ablation          & $58\% \pm 28.2\% $ \\
  CAP                             & $96\% \pm 3.5\% $ \\
  \bottomrule
\end{tabular}
\label{tab:ablation-close}
\end{table}
To ablate the contact anchor, we choose the Close task since the task objective (an open door or drawer) is visually apparent without ambiguity, even without the contact anchor conditioning. We train an RGB-only ablation of \capshort{}, and run the same evaluation procedure as described in \cref{ssec:open-close-eval-main}. While the vanilla \capshort{} achieved 96\% on the Close task, the RGB-only ablation performed much worse at 58\% (\cref{tab:ablation-close}), validating our hypothesis that having a contact anchor improves manipulation performance.

\subsubsection{Distractor objects and policy performance}
\begin{figure}[h!]
    \centering
    \includegraphics[width=0.5\linewidth]{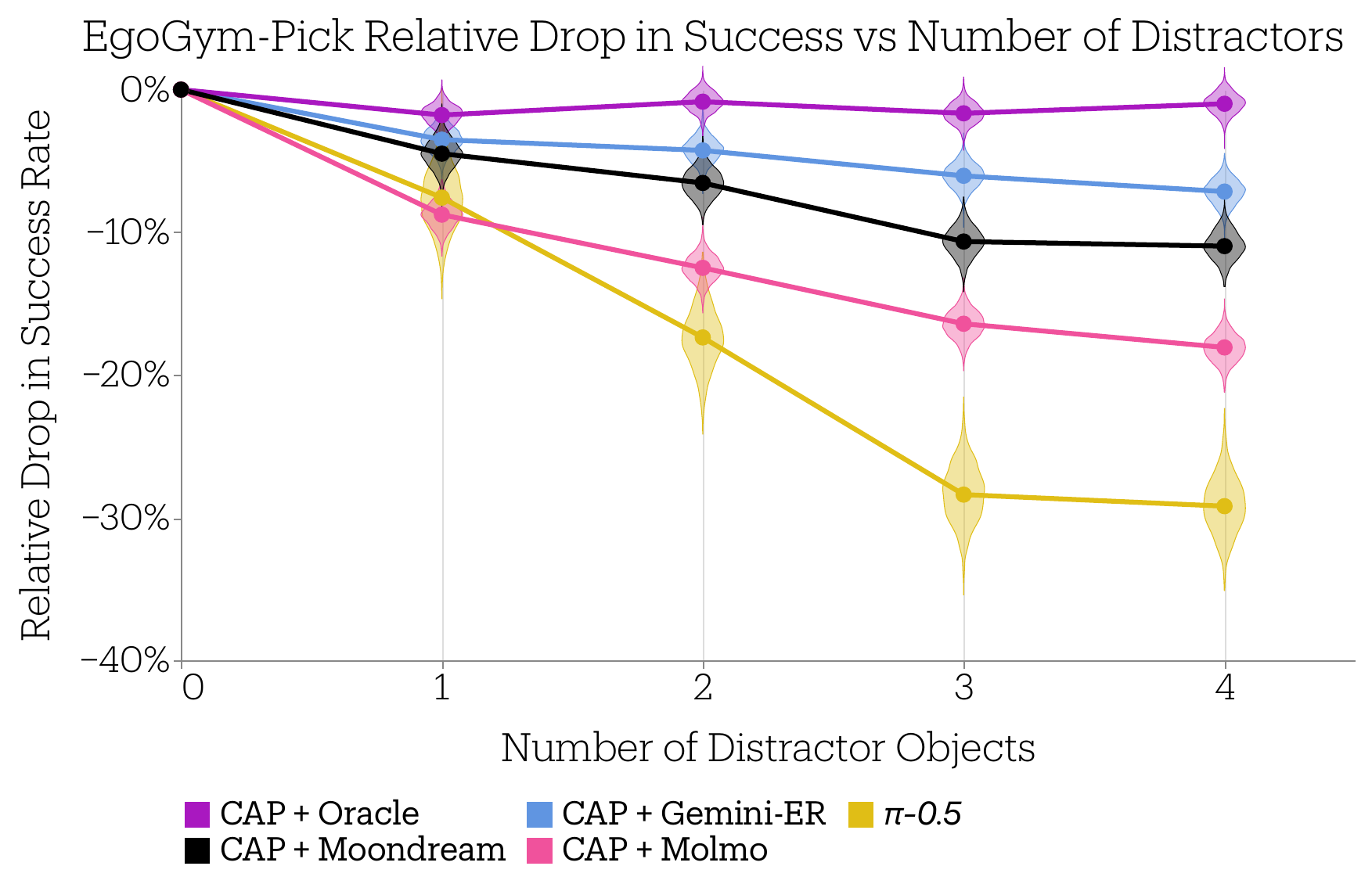}
    \caption{Relative success rate as a function of visual distractors for CAP and $\pi_{0.5}$ models on \gym{}-Pick. Success rates are normalized to each model's performance with zero distractors.}
    \label{fig:distractors}
\end{figure}

We evaluate \capshort{} with various VLMs for contact prompt generation, as well as $\pi_{0.5\texttt{\small-DROID}}$ in the \gym{}-Pick environment. The baseline \capshort{} with privileged oracle contact anchor information remains stable in performance regardless of distractor objects. We see a decrease in performance for \capshort{} + VLMs and $\pi_{0.5}$, picking up more wrong objects as we increase the number of distractors in the scene.

\section{Related Works}

\subsection{Generalist Behavior Models}
Current progress in learning-based robotic systems and advent of large-scale data in robotics has enabled development of general robot manipulation policies: models that can be applied to novel scenes, objects, tasks, or robots without having trained on the same~\citep{etukuru2024rum,Black2024Pi0,PI2025Pi05,hu2024data}. In most cases, the standard recipe requires creating and training on a large training set which contains diversity in the intended axes of generalization. Note that this paradigm is different than many large robotic models in literature~\citep{team2024octo,tri2025lbm,NVIDIA2025GR00T,Lee2025MolmoAct,shukor2025smolvla} which do not claim zero-shot performance and requires some post-training in the particular evaluation setup (task, scene, or robot) to show reasonable behavior in that setup.

Current understanding on the necessary and sufficient amount of scale in data, model size, or compute for such generalization is still in its infancy. Existing multi-task generalist models, mostly proprietary, uses at least 1,000 to 10,000 hours of data, while single-task general policies such as~\citep{etukuru2024rum, hu2024data, chi2024universal} has been trained on as little as 1,000 demonstrations per task.

\subsection{Conditioning Multi-modal Behavior Models}
A general policy capable of multiple possible behaviors in a particular environment needs to be conditioned by user intent or goal to elicit useful behavior. Earliest forms of conditioning behavior policies relied on communicating a future state or image~\citep{lynch2020learning,cui2022play,bousmalis2023robocat} to the robot. With the rise of capable language models, language became a popular mode for conditioning language starting with~\citep{lynch2020grounding,jang2021bc,brohan2022rt,brohan2023rt2}. In these works, language is used directly as an input modality to the model. However, advent of multimodal grounding models such as~\citep{radford2021clip} allowed~\citep{shridhar2022cliport,shafiullah2022clip,shridhar2022perceiver} to ground language in some spatial conditioning as an input to the robot. Other low-dimensional policy conditioning include~\citep{gu2023rt,sundaresan2024rt} where robot tranjectory and a sketch of the goal is used to condition the model. The most related to our work is RT-trajectory~\citep{gu2023rt}, where gripper motion and articulation hindsight relabelling was used to condition a model. Our method simplifies the premise to focus only on the contacts between the robot and the environment as the minimal interface, and shows that it already leads to strong general behavior models.
Another line of work uses keypoints extracted by pretrained models to give robots generalization abilities~\citep{haldar2025point,levy2025p3,huang2024rekep,bharadhwaj2024track2act}. However, they forgo the pixel-to-action paradigm by using only keypoints as observation or using a planner to generate robot actions.

\subsection{Evaluating Real Manipulation Policies in Simulation}
One of the biggest bottlenecks in training policies for real robot is evaluation~\citep{zhou2023train}: training or test losses are unindicative of policy success, and getting statistically significant effect sizes require onerous evaluation schedules~\citep{tri2025lbm}. Therefore, there has been significant interest in using simulated setups to evalute real robot performance~\citep{tri2025lbm,jain2025polaris,li2024evaluating,jangir2025robotarena}. A shared focus of these works is to start from some ground-truth real world environments and then model them as closely as possible in simulations. This focus on simulation fidelity purportedly serves to ensure transfer of simulated evaluation results to real world. But even minor differences between sim and real can make it difficult to hill-climb a simulation metric that will lead to real world improvement~\citep{tri2025lbm}. Note that, these simulation benchmarks have a different goal than~\citep{liu2024libero,li2024behavior,Nasiriany2024RoboCasa} where the goal is to compare different learning algorithms trained on a fixed set of data and not general task competence.

In this work, we instead look at factored, single task simulation with many procedurally generated scenes, which makes overfitting to this metric difficult. Indoor navigation has successfully used this approach~\citep{eftekhar2024one} through simulations providing a large number of household environments~\citep{Kolve2017AI2THOR,Savva2019Habitat,Deitke2022ProcTHOR}.

\section{Conclusion}
\label{sec:conclusion}

In this work, we introduce~\caplong{} (\capshort{}), a principled way of conditioning general policies that achieves superior zero-shot performance on single tasks with a modestly sized dataset, model size, and compute budget. We additionally demonstrate how we can develop such~\capshort{}s with simulation in the loop, and also chain together~\capshort{}s via tool calling to perform long-horizon manipulations. With~\capshort{}, we hope to provide the right framework for researchers with limited resources, such as those in academia, to study the emergence of general behavior in robotics.
While~\capshort{} makes advances in the problem of learning general manipulation policies, we are only able to scratch the surface and there are many aspects of~\capshort{} that we think should be studied in future work. Firstly, extending~\capshort{} to tasks with multiple contact anchors or for bimanual tasks would be useful, and will require extending the system to predict and ingest multiple contacts or even a distribution of them. Secondly,~\capshort{}s rely on two input modalities, and studying the process through which they decide on their relative weights in the decision making may elucidate fundamental factors about the dynamics of supervised policy learning itself. Finally, as a practical matter, exploring how we could roll the verifier guided retrying process into the end-to-end part of the policy via either real world or simulated reinforcement learning would go a long way into making~\capshort{}s practical for high-stakes applications.

\section*{Acknowledgments}
We thank Shenglong Wang and the NYU HPC team for helping us with compute,
Kanad Patel and Dhawal Kabra for helping with the robot hardware, Gaoyue Zhou, Raunaq Bhirangi, Irmak Güzey, and Nikhil Bhattasali for insightful discussions, and Andrew Liao, Lin Yang, Peng Jiang, and Santosh Srinivas for contributions to the dataset.
NYU authors are supported by grants from Honda, Hyundai, NSF award 2339096 and ONR awards N00014-21-1-2758 and N00014-22-1-2773. LP is supported by the Packard Fellowship and Sloan Fellowship. Hello Robot authors are supported by NIH NIA R43AG072982.

\bibliographystyle{yearnat}
\bibliography{references_with_urls}

\clearpage
\appendix
\section{Appendix}

\subsection{Data collection details}
\label{app:sec:data_collection}

\subsubsection{Visual gripper state estimation by SAM2}
\begin{figure}[ht!]
    \centering
    \includegraphics[width=0.35\linewidth]{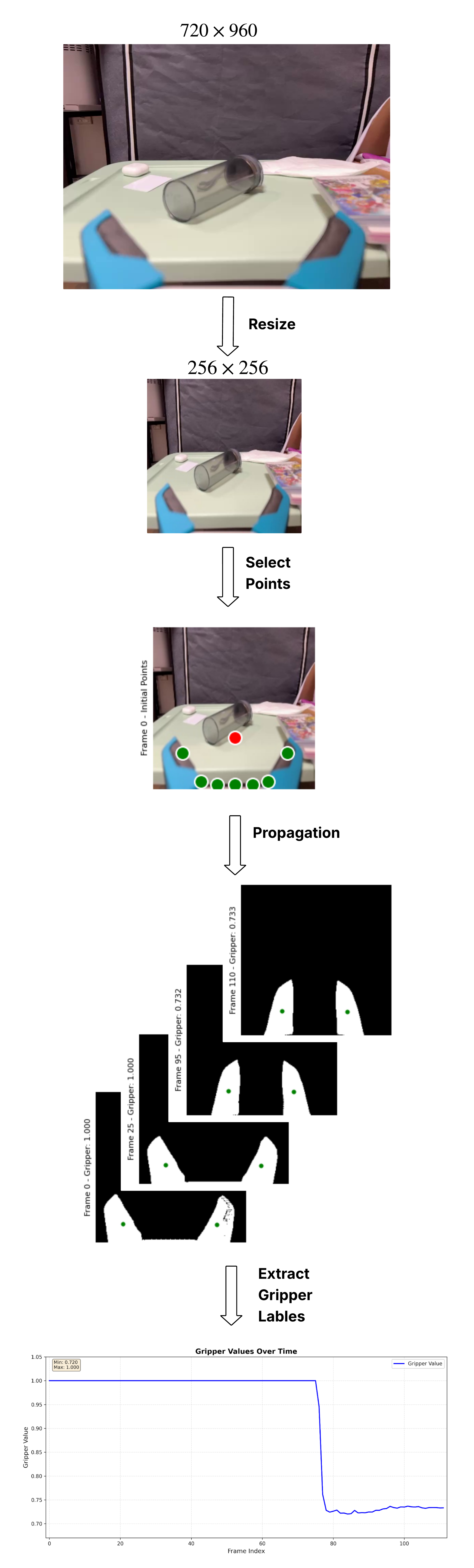}
    \caption{Pipeline for Extracting Gripper Label By Using SAM2}
    \label{fig:pipeSam2}
\end{figure}

We resize the video observation to $256 \times 256$. At the start of each video, the gripper is fully open. We prompt SAM2 with positive points (green) belonging to the gripper, and negative point (red) outside the gripper, shown in \cref{fig:pipeSam2}, to generate a gripper segmentation mask for all frames. We compute the center of mass of the left and right gripper from the binary mask. The horizontal pixel distance between these two
centers represents the gripper aperture, linearly mapped to 0 (fully closed) to 1 (fully open).

\subsubsection{Data Filtering and Augmentation}

\paragraph{Static Frame Filtering}
We filter out frames where the gripper does not significantly move. Starting from the first frame, we scan forward and select a new frame when the cumulative movement exceeds 0.3cm in translation, 0.1 radians in rotation, or 0.05 in gripper aperture.

\paragraph{Trajectory Mirroring}
We apply horizontal flips on each recorded demo for both Open and Close tasks. We keep the original trajectory, and a horizontally mirrored copy of the trajectory with flipped visual observations and end effector poses.

\subsection{Policy training details}
\label{app:sec:policy_training}

\subsubsection{Visual encoder pretraining}

We pretrain ResNet-50 backbones with MoCo \citep{he2019momentum} on recorded RGB observations. The Pick encoder is trained on frames from 14,606 demonstrations (16.3 hours) across 289 environments, while the Open/Close encoder is trained on frames from 5,759 demonstrations (6.7 hours) across 135 environments.

\begin{table}[ht]
\caption{Hyperparameters used for Pick, Open, and Close tasks}
\label{tab:hyperparams}
\centering
\begin{tabular}{lccc}
\hline
\textbf{Hyperparameter} & \textbf{Pick} & \textbf{Open} & \textbf{Close} \\
\hline
Obs window size & 3 & 3 & 3 \\
\hline
Training Steps & 308,565 & 364,811 & 277,522 \\
Batch Size & 256 & 200 & 200 \\
Learning Rate & 3e-4 & 1e-4 & 2.7e-4 \\
\hline
Transformer Depth & 8 & 8 & 8 \\
Attn Heads & 8 & 8 & 8 \\
Embedding Dim & 512 & 512 & 256 \\
\hline
VQ-VAE codebook size & 16 & 32 & 32 \\
VQ-VAE embedding dim & 512 & 512 & 512 \\
\hline
\end{tabular}
\end{table}

\subsection{Policy deployment details}
\label{app:sec:deployment}
\begin{figure}[ht!]
    \centering
    \includegraphics[width=\linewidth]{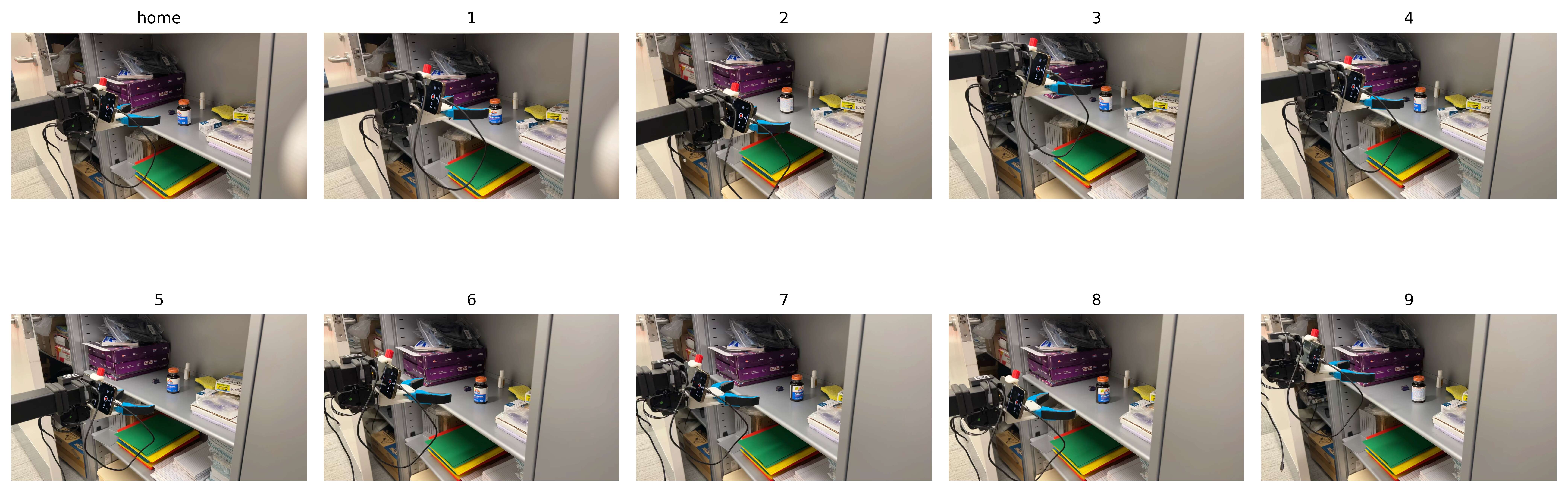}
    \caption{Real robot configurations corresponding to each starting pose}
    \label{fig:robot_starting_poses_images}
\end{figure}
\begin{figure}[ht!]
    \centering
    \includegraphics[width=0.5\linewidth]{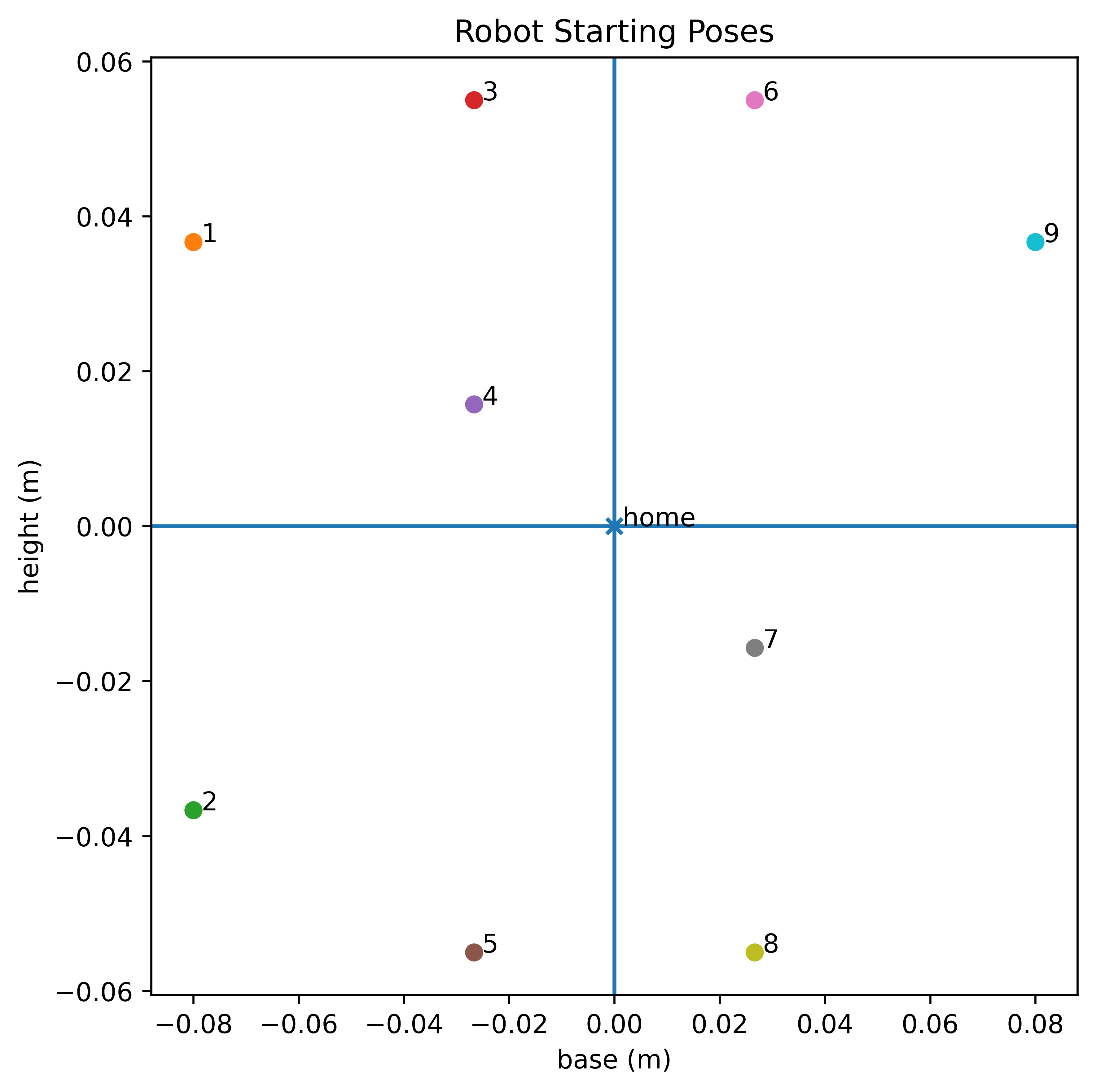}
    \caption{Robot starting poses in the base--height parameter space}
    \label{fig:robot_starting_poses_scatter}
\end{figure}

For policy deployment on Stretch, we run inference at up to 2Hz directly on CPU on the onboard Intel NUC. For deployment on xArm, Franka, and other fixed embodiments, we run inference on an NVIDIA RTX A4000 GPU.

All evaluation objects and environments used in Pick, Open, and Close are unseen during training. See Figure \cref{fig:pick-eval-objects} for the 25 objects used Pick evaluation.

For each object used in the Pick and each door and drawer used in the Open and Close, we have 10 trials. Correspondingly, we start the robot at 10 initial positions depicted in  \cref{fig:robot_starting_poses_scatter}, \cref{fig:robot_starting_poses_images}.

\subsubsection{Evaluation object details}
For the Pick task, see \cref{fig:pick-eval-objects} for a display of the 25 objects used for evaluation. For the Open and Close tasks, see \cref{fig:eval_envs} for a display of the five doors and five drawers used for evaluation.

\begin{figure}[b!]
    \centering
    \includegraphics[width=\linewidth]{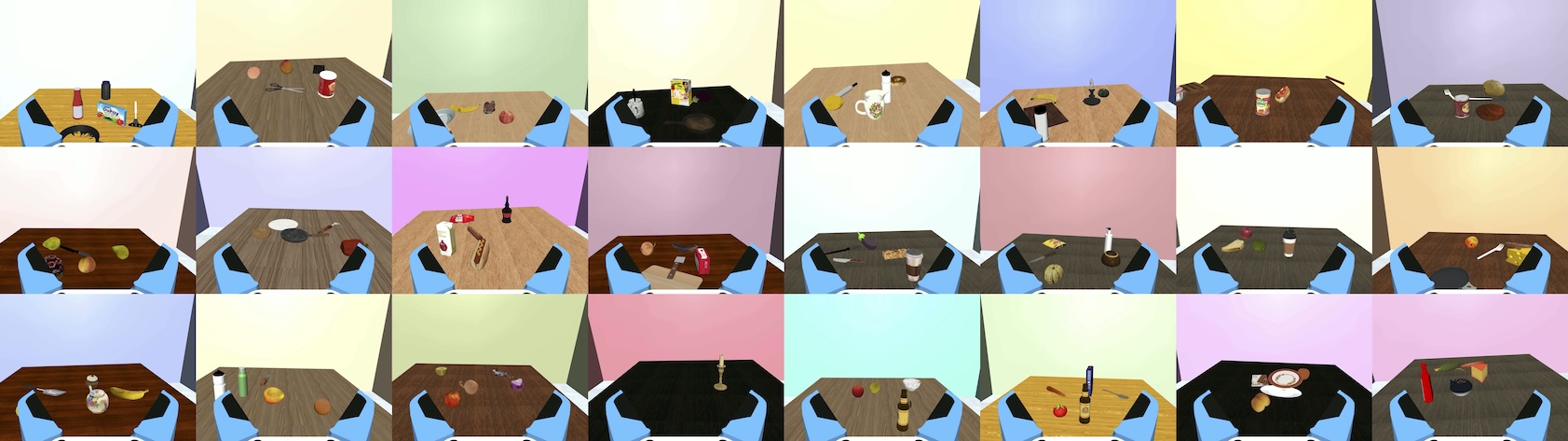}
    \caption{EgoGym Pick environment visualizations}
    \label{fig:egogym-pick-eval}
\end{figure}

\begin{figure}[b!]
    \centering
    \includegraphics[width=0.6\linewidth]{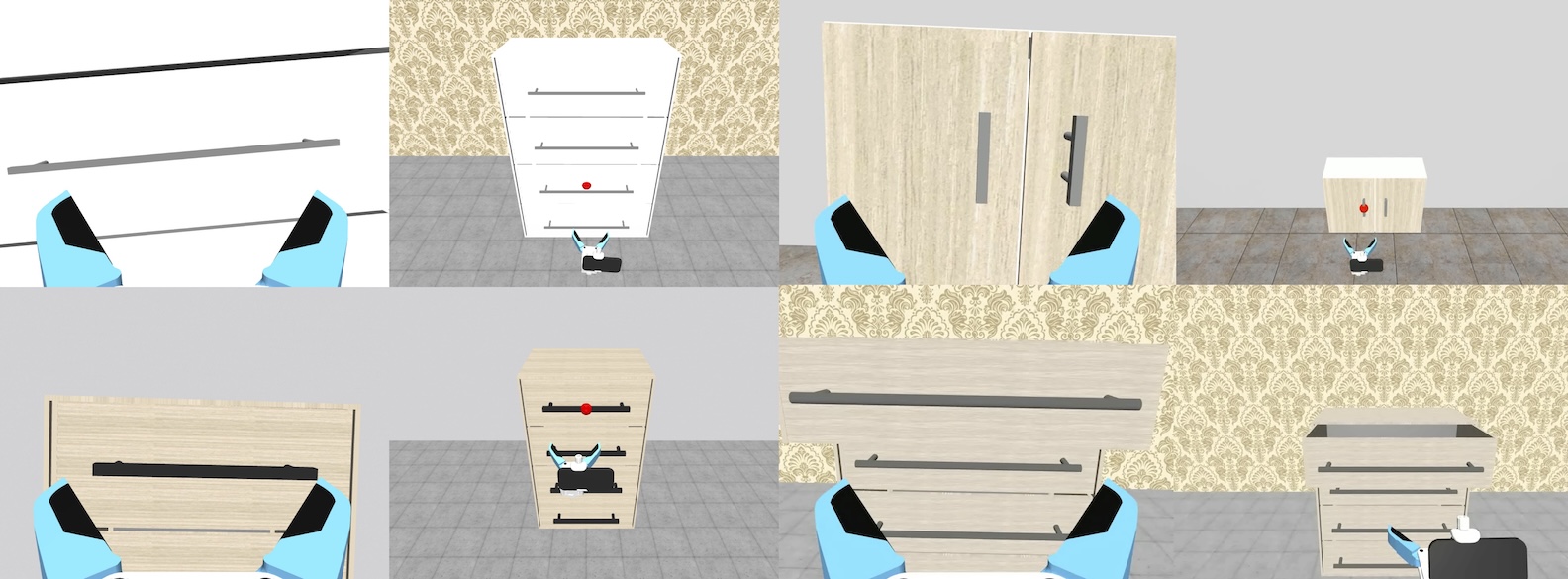}
    \caption{EgoGym Open/Close environment visualizations}
    \label{fig:egogym-open-close}
\end{figure}

\subsection{Simulation: \gym ~details}
\label{app:sec:simulation}

\paragraph{Available environments}
EgoGym provides three Gymnasium environments:
\begin{itemize}
    \item EgoGym-Pick-v0: a tabletop pick-and-lift task.
    \item EgoGym-Open-v0: opening an articulated object.
    \item EgoGym-Close-v0: closing an articulated object.
\end{itemize}

Figures~\ref{fig:egogym-pick-eval} and~\ref{fig:egogym-open-close} show visualizations of the EgoGym Pick and Open/Close environments.

\paragraph{Environment initialization arguments}

Each environment is configurable through a set of arguments. The robot embodiment can be either \emph{CAP} or \emph{DROID}, and actions can be (\emph{relative}) or (\emph{absolute}). The scene is populated by sampling from a specified object set.

Optionally, environments can be wrapped with a VLM to provide unprivileged perception. Supported models include Moondream \citep{vik_2024}, Gemini-Robotics-ER-1.5 \citep{team2025gemini}, and Molmo \citep{deitke2024molmo}. When no wrapper is used, the policy has privileged access to object identity.

\paragraph{Observations}
All tasks expose a shared visual and proprioceptive observations. For the DROID embodiment, joint positions are additionally provided. In the Pick task, the observation includes the pose of the target object. In the Open and Close tasks, the observation instead includes the pose of the object's handle.

\paragraph{Rewards}
Each environment provides a simple dense reward signal. In the Pick task, reward is given by the vertical displacement of the target object relative to its initial placement. In the Open task, reward corresponds to the fraction of the articulated object that has been opened. In the Close task, reward is computed as a residual-to-closed score from the object's opening percentage.

\paragraph{VLM Distractor evaluation protocol}
For the Pick task, we evaluate robustness to distractors by varying the number of objects in the scene from one to five, across different VLMs. Evaluation is performed over 5{,}000 episodes. Episodes are considered successful if the maximum reward exceeds a threshold of $0.03$. All experiments use a maximum horizon of 80 steps.

\paragraph{$\pi_{0.5}$ baseline (Pick)}
We additionally evaluate the $\pi_{0.5}$ baseline on EgoGym-Pick-v0 using the pi05\_droid\_jointpos checkpoint hosted at \texttt{gs://openpi-assets/checkpoints}. These experiments use a task horizon of 350 steps. Evaluation is performed over 5{,}000 episodes with the same success threshold as above.

\paragraph{Failure mode classification (Pick).}
We assign each episode to exactly one outcome using an ordered decision tree (first match wins):

\begin{enumerate}
    \item \textbf{Success}: Maximum lift $> 0.03$m.

    \item \textbf{Did not lift enough}: The gripper fingers contacted the target object at least once and maximum lift $\geq 0.005$m ($\geq 0.5$cm), but not successful.

    \item \textbf{Object touched but not grasped}: The gripper fingers contacted the target object at least once, but did not lift.

    \item \textbf{Picked wrong object}: The gripper fingers contacted a non-target object at least once, but never contacted the target object (i.e., only distractor contact).

    \item \textbf{Empty Grasp}: None of the rules above triggered, but the episode ends with the gripper in a closed state or the gripper fingers made contact with each other, indicating a close-on-nothing.

    \item \textbf{Did not grasp}: None of the above (no meaningful contacts).
\end{enumerate}

\begin{figure}[h!]
    \centering
    \begin{minipage}{0.48\linewidth}
        \centering
        \includegraphics[width=\linewidth]{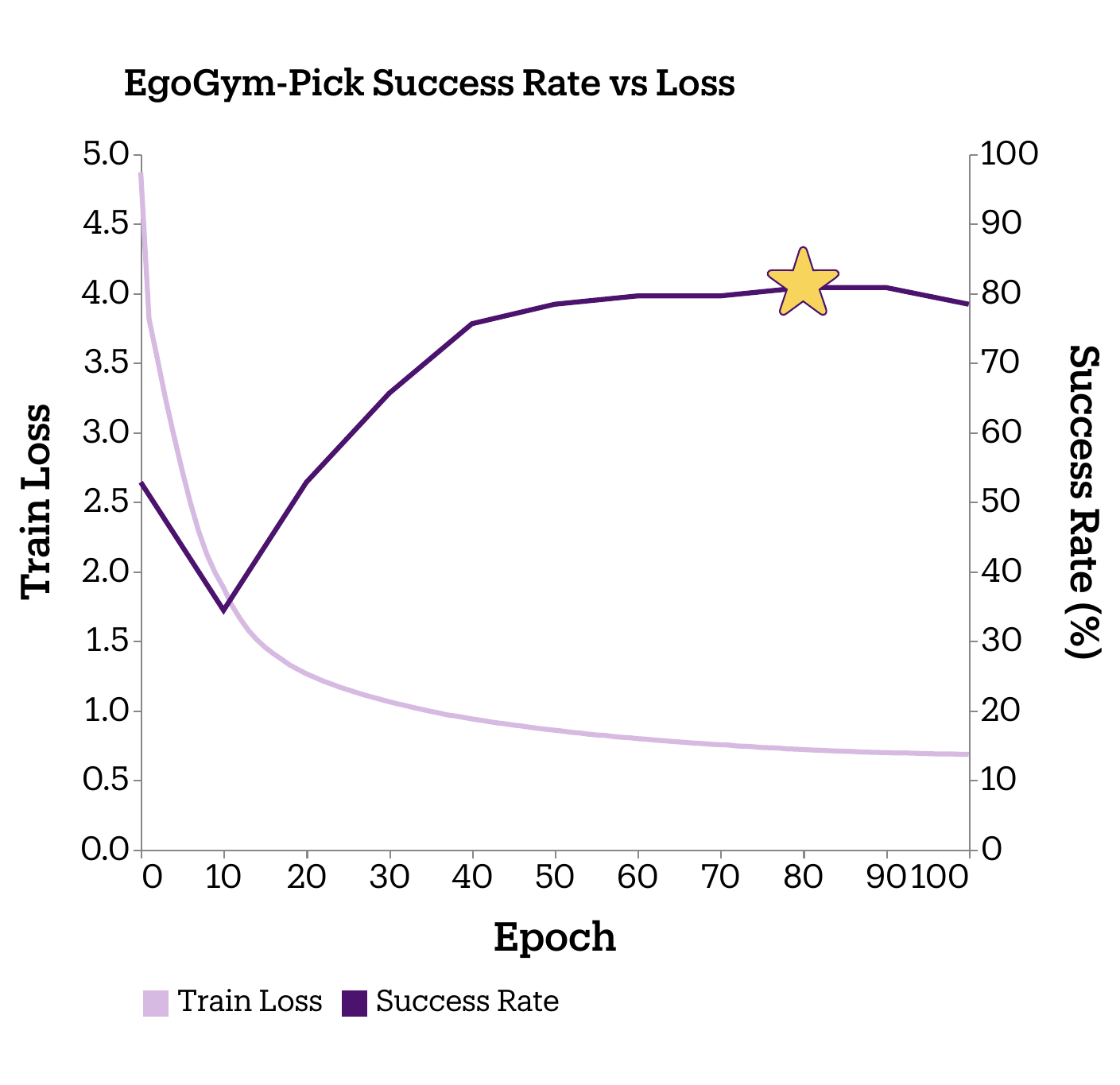}
        \caption{Loss and simulation success rate over time for the Pick task during a CAP training run.}
        \label{fig:loss_vs_sim_sr_pick}
    \end{minipage}
    \hfill
    \begin{minipage}{0.48\linewidth}
        \centering
        \includegraphics[width=\linewidth]{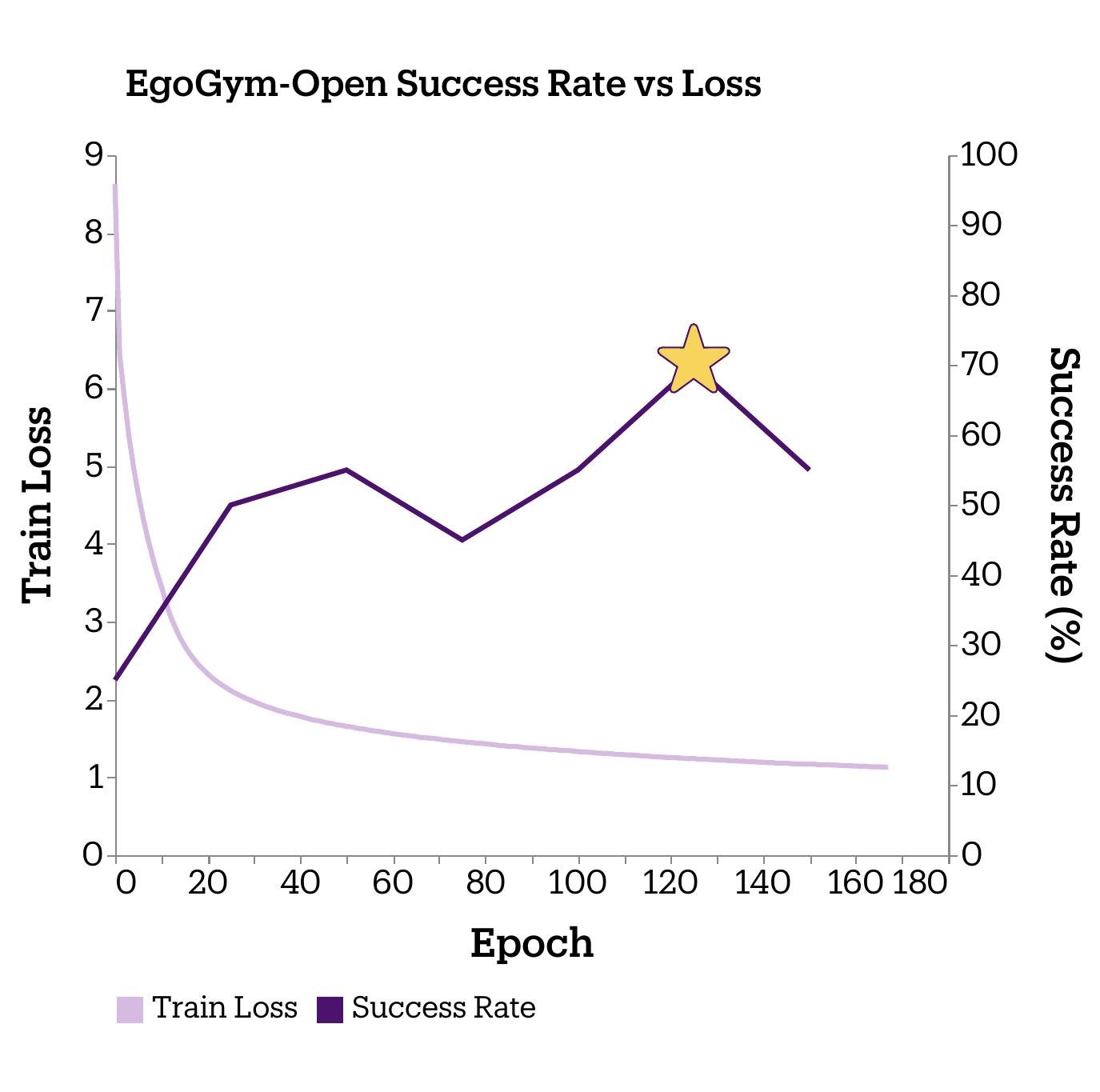}
        \caption{Training loss and simulation success rate over time for the Open task during a CAP training run.}
        \label{fig:loss_vs_sim_sr_open}
    \end{minipage}
\end{figure}

\begin{figure*}[t!]
    \centering
    \includegraphics[width=\linewidth]{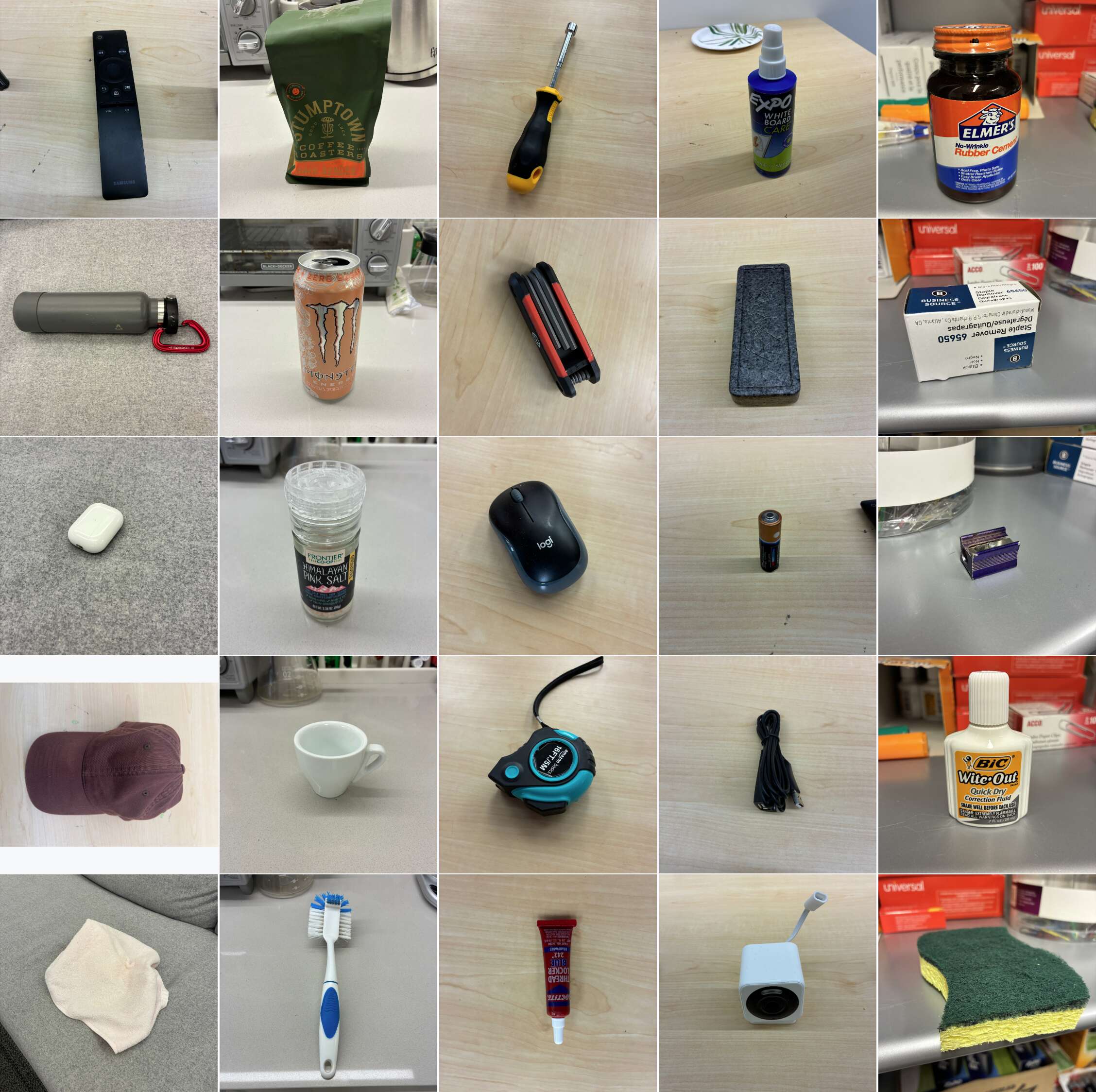}
    \caption{Evaluation objects used for the Pick evaluations}
    \label{fig:pick-eval-objects}
\end{figure*}

\paragraph{Integration into the training loop}
EgoGym is integrated into the training loop to periodically evaluate task success rates, since loss alone is not a reliable indicator of policy performance. Figures~\ref{fig:loss_vs_sim_sr_pick} and~\ref{fig:loss_vs_sim_sr_open} show example CAP training runs, reporting simulation success rate alongside loss for the Pick and Open tasks, respectively.

\label{app:sec:eval_objects}

\end{document}